\documentclass[11pt]{article}
\usepackage{graphicx}
\usepackage{xr-hyper}
\usepackage{amsmath}
\usepackage{upgreek}
\usepackage{amssymb}
\usepackage{amsfonts}
\usepackage{amsthm}
\usepackage{mathrsfs}
\usepackage{fontenc}
\usepackage{empheq}
\usepackage{enumerate}
\usepackage{comment}
\usepackage{appendix}
\usepackage{mathtools}
\usepackage[shortlabels]{enumitem}
\usepackage{algorithm}
\usepackage[ruled,algo2e]{algorithm2e}
\usepackage{color}
\usepackage{authblk}
\usepackage{url}

\makeatletter
\DeclareMathOperator*{\argmin}{arg\,min} 
\DeclareMathOperator*{\argmax}{arg\,max} 

\mathtoolsset{showonlyrefs}
\theoremstyle{plain}
\newtheorem{thm}{Theorem}
\newtheorem{lem}{Lemma}
\newtheorem{prop}[thm]{Proposition}

\newtheorem{assu}{Assumption}

\theoremstyle{definition}

\theoremstyle{remark}
\newtheorem{rem}{Remark}

\newcommand{\bR}{\mathbb {R}}
\newcommand{\bH}{H}
\newcommand{\bP}{\mathbb {P}}
\newcommand{\bS}{\mathcal {S}}
\newcommand{\bA}{\mathcal {A}}
\newcommand{\EE}{\mathbb {E}}
\newcommand{\cond}{\,|\,}

\newcommand{\rmd}{\,\mathrm{d}}

\begin{document}

\title{An $L^2$ Analysis of Reinforcement Learning in High Dimensions with Kernel and Neural Network Approximation}

\author[1]{Jihao Long}
\author[2,3]{Jiequn Han\thanks{Corresponding author. Email addresses: jihaol@princeton.edu (J. Long), jiequnhan@gmail.com (J. Han), weinan@math.princeton.edu (W. E)}}
\author[4,2,1]{Weinan E} 
\affil[1]{Program of Applied and Computational Mathematics,
          Princeton University}
\affil[2]{Department of Mathematics, Princeton University}
\affil[3]{Center for Computational Mathematics, Flatiron Institute}
\affil[4]{School of Mathematical Sciences, Peking University}

\date{}

\maketitle
\begin{abstract}
Reinforcement learning (RL) algorithms based on high-dimensional function approximation have achieved tremendous empirical success in large-scale problems with an enormous number of states. However, most analysis of such algorithms gives rise to error bounds that involve either the number of states or the number of features. This paper considers the situation where the function approximation is made either using the kernel method or the two-layer neural network model, in the context of a fitted Q-iteration algorithm with explicit regularization. We  establish an $\tilde{O}(H^3|\bA|^{\frac14}n^{-\frac14})$ bound for the optimal policy with $Hn$ samples, where $H$ is the length of each episode and $|\bA|$ is the size of action space. Our analysis hinges on analyzing the $L^2$ error of the approximated Q-function using $n$ data points. Even though this result still requires a finite-sized action space,  the error bound is independent of the dimensionality of the state space.  
\end{abstract}

\section{Introduction}

Modern reinforcement learning (RL) algorithms often deal with problems involving an enormous amount of states, often in high dimensions, where function approximation must be introduced for the value or policy functions. Despite their practical success~\cite{mnih2013playing,silver2016mastering,duan2016benchmarking}, most existing theoretical analysis of RL is only applicable to the tabular setting~(see e.g.~~\cite{jaksch2010near,azar2012sample,osband2016generalization,azar2017minimax,dann2017unifying,jin2018q}), in which both the state and action spaces are discrete and finite,  and the value function is represented by a table without function approximation. Relatively simple function approximation methods, such as 
the linear model~\cite{yang2019sample,jin2020provably} or generalized linear model~~\cite{wang2019optimism}, have been studied in the context of RL with various statistical estimates. The kernel method has also been studied in ~~\cite{farahmand2016regularized,domingues2020regret,yang2020provably,yang2020function}, but results therein either suffer from the curse of dimensionality or require stringent assumptions about the kernel (in the form of fast decay of the kernel's eigenvalues or the bounds on the covering number). 
This paper considers general kernel method and two-layer neural network models and establishes dimension-independent results for these two classes of function approximation.

In the context of supervised learning,  dimension-independent error rates have been established for a number of important
machine learning models~\cite{weinan2020towards}, including the kernel methods and two-layer neural network models.
Of particular importance is the choice of the function space associated with the specific machine learning model.
For the kernel method and two-layer neural network models, the corresponding function spaces are the reproducing kernel Hilbert space (RKHS)~\cite{aronszajn1950theory} and Barron space~\cite{weinan2019barron}, respectively. 
Extending such results to the setting of reinforcement learning  is a challenging task due to the coupling between the value/policy functions at different time steps.

In this work, we consider the fitted Q-iteration algorithm~\cite{ernst2005tree,riedmiller2005neural,szepesvari2010algorithms,chen2019information,fan2020theoretical} 
for situations where the state space is embedded in a high-dimensional Euclidean space and the action space is finite. 
The function approximation to the Q-function is made either using the kernel method or the two-layer neural networks, with explicit regularization.
We assume there is a \textit{simulator} that can generate samples for the next state and reward, given the current state and the action. 
This allows us to focus on analyzing errors from the function approximation. 
Under the assumptions that the function approximation is compatible with the reward function and transition model, and all admissible distributions are uniformly bounded from a reference distribution (see Assumption \ref{assumption_concentration}), we establish an $\tilde{O}(H^3|\bA|^{\frac14}n^{-\frac14})$ bound for the optimal policy with $Hn$ samples, where $H$ is the length of each episode and $|\bA|$ is the size of action space. 
This result is independent of the 
dimensionality of the state space, and the convergence rate for $n$ is close to the statistical lower bound for many function spaces, including several popular cases of RKHS and the Barron space (see Section \ref{section_sample} for a detailed discussion).
The key component in the analysis is to estimate the one-step error and control the error propagation.
An important issue is the choice of the norm.
$L^\infty$ estimates have been popular in reinforcement learning for analyzing the tabular setting \cite{jaksch2010near,azar2017minimax,dann2017unifying,jin2018q}, linear models, \cite{cai2019provably,yang2019sample,jin2020provably} and kernel methods \cite{yang2020provably,yang2020function} (see discussions below).
However, in the case we are considering,  %
$L^\infty$ estimates suffer from the curse of dimensionality with respect to the sample complexity, 
i.e. to ensure that the error is smaller than  $\epsilon$, we need at least $O(\epsilon^{-d})$ samples in $d$-dimensional state space (see Section~\ref{section_sample} for a detailed discussion). 
This fact also explains why we only consider finite action space. Once we consider the high-dimensional action space, it is inevitable to find a maximum of a high-dimensional function based on finite samples, which also suffers the curse of dimensionality.
In contrast, we choose to work with the $L^2$ estimates.
$L^2$ estimates have been a successful tool in supervised learning  and led to dimensionality-independent error rates for kernel methods \cite{cortes2010generalization,cortes2013learning} and neural network models \cite{ma2019priori}. 
In reinforcement learning, $L^2$ estimates have been used  in the case with a finite set of candidate approximating functions~\cite{chen2019information}, 
composition of H\"older functions with certain sparse structures~\cite{fan2020theoretical}, or for kernel methods under the additional assumption on the covering number of the unit ball in RKHS~\cite{farahmand2016regularized} (see discussions below).
 To deal with error propagation,  certain assumptions are needed on the concentration coefficients (see Assumption \ref{assumption_concentration} below).

\textbf{Related Literature.} In addition to the results discussed above, error estimates and sample efficiency in conjunction with function approximation have also been studied in the setting when the transition dynamics is fully deterministic~\cite{wen2013efficient} or 
has low variance~\cite{du2019provably}. 
In the setting of episodic reinforcement learning where the agent can only have whole trajectory data, several works have studied efficient exploration in the state space based on different function approximations~(see e.g.~\cite{domingues2020regret,zanette2020frequentist,yang2020reinforcement}).
For policy based algorithms,  the convergence of the policy gradient methods has been analyzed in \cite{cai2019provably,cai2019neural,liu2019neural,wang2019neural}  with linear function approximation or in the highly over-parameterized regime under assumptions similar to Assumption \ref{assumption_concentration} using the theory of neural tangent kernel \cite{jacot2018neural}. It should be noted that in the neural tangent kernel regime,
the two-layer neural network model only generalizes well for target functions in an associated RKHS with
kernel defined by the initialization \cite{jacot2018neural,neyshabur2018towards,arora2019fine,ma2019comparative}.  This is a much smaller space than the Barron space that we use in this paper.
\cite{farahmand2016regularized} considers the kernel method in the time-homogeneous setting and proves that we can obtain an $\tilde{O}((1-\gamma)^{-2}(n^{-\frac{1}{2(1+\alpha)}} +\gamma^K))$ bound for the optimal policy with $Kn$ samples where $0 < \gamma < 1$ is the discount factor and $K$ is the number of iterations. This result builds on the assumption that the covering number satisfies
\begin{equation}
    \log \mathcal{N}_\infty(u, \mathcal{F}) \le C u^{-2\alpha},
\end{equation}
for any $u > 0$.  Here $C > 0$ and $0 < \alpha < 1$ are two constants and $\mathcal{F}$ is the unit ball in the corresponding RKHS 
(see Assumption A4 in \cite{farahmand2016regularized} for a detailed discussion). Note that
existing results on the covering number of the unit ball do not apply to general RKHS ~\cite{zhou2002covering,zhou2003capacity,steinwart2008support,yang2020function}.
\cite{yang2020provably,yang2020function} also consider the kernel method. Their results rely on the decay of the kernel's eigenvalues. 
Assuming that the $j$-th eigenvalue (see Section~\ref{section_sample}  for a formal definition) satisfies $\lambda_j \ge C j^{-\gamma}$ for any $j \ge 1$ 
with positive constants $C > 0$ and $\gamma > 1$, then 
the total regret~\cite{bubeck2012regret} in $T$ episodes satisfies 
$$
    \mathrm{Regret}(T) \le \tilde{O}(H^2 T^{\kappa^{*} + \xi^{*}+1/2}),
$$
where $\kappa^{*}$ and $\xi^{*}$ are two parameters depending on dimension $d$ and $\gamma$. 
From this a dimension-independent estimate follows when $\gamma$ is sufficiently large.
However, when $\gamma \le d+2$ the right-hand side becomes $O(T)$ and hence the bound
becomes trivial. See Section~\ref{section_sample} for more discussions. 
In contrast, our algorithm is applicable to general RKHS and we  obtain an $\tilde{O}(H^3|\bA|^{\frac{1}{4}} n^{-\frac{1}{4}})$
 bound for the optimal policy with $Hn$ samples for this algorithm.

\textbf{Notation.}  Let $\bS$ be a subset of Euclidean space $\bR^d$. We use $C(\bS)$ and $\mathcal{P}(\bS)$ to denote the continuous function space on $\bS$ and the collection of all probability distributions on $\bS$ respectively. $\|\cdot \|_{p,\mu}$ denotes the $L^p$ norm under the probability measure $\mu \in \mathcal{P}(\bS)$. By default we use $\mathcal{F}$ to denote a function class
\begin{align}
    \{f: \bS \times \bA \mapsto \mathbb{R} \cond f(\,\cdot, a)\in C(\bS) \text{ for any } a\in \bA\}.
\end{align}
$x\cdot y$ denotes inner product in the Euclidean space.
$\mathbb{S}^{d-1}$ denotes the unit $(d-1)$-sphere: $\{x\in \mathbb{R}^{d}:\, \|x\|=1\}$.
$[H] = \{1,\dots, H\}$. For any $m > 0$,
we define the truncation operator:
\begin{equation}\label{truncated_operator}
\mathcal{K}_mf(x) = \min\{ \max\{f(x),0\},m\}. 
\end{equation}
We use standard big O notation $O(\cdot)$ that hides absolute constants and $\tilde{O}(\cdot)$ that hides both absolute constants and poly-logarithmic factors.

\section{Background}
In this section, we introduce the background. We first lay out the formulation of Markov decision process in the reinforcement learning problem. We then introduce two function approximation approaches considered in this paper: kernel method and two-layer neural network model. In both cases, we state our assumptions on the reward function and transition probabilities in terms of function norms in the corresponding function spaces.
\subsection{Markov Decision Processes}
We consider an episodic Markov decision process (MDP) $(\bS,\bA,H,\bP,r)$. Here $\bS$ denotes the set of all the states, which is a subset of Euclidean space $\bR^d$, and $\bA$ denotes the set of all the actions. In this paper, we are particularly interested in the case where the state's dimension $d$ is large and $\bA$ is a finite set with cardinality $|\bA|$.
$H$ is the length of each episode, $\bP = \{\bP_h\}_{h=1}^H$ and $r = \{r_h\}_{h=1}^H$ are the state transition probability measures and reward functions, respectively. For each $h \in [H]$, $\bP_h(\,\cdot \cond x,a)$ denotes the transition probability for the next state at the step $h$ if current state is $x$ and action $a$ is taken. $r_h: \bS \times \bA \mapsto [0,1]$ is the deterministic reward function at step $h$. We choose to work with deterministic reward  to avoid complicated notations. Our analysis can be easily generalized to the setting where the reward is stochastic. 

We denote a policy by $\pi = \{\pi_h\}_{h=1}^H \in \mathcal{P}(\bA \cond \bS)$, where
\begin{align}
    \mathcal{P}(\bA\,|\,\bS) = \Big\{\{\pi_h(\,\cdot \cond \cdot\,)\}_{h=1}^H,\pi_h(\,\cdot \cond x) \in \mathcal{P}(\bA)
    \text{ for any }x \in \bS \text{ and } h \in [H]\Big\}.
\end{align}
Given an initial state distribution $\mu \in \mathcal{P}(\bS)$, we define the total reward function for any $\pi \in \mathcal{P}(\bA \cond \bS)$ as follows
\begin{equation}
 J_\mu(\pi) = \EE_{\mu,\pi}[\sum_{h=1}^H r_h(S_h,A_h)],
\end{equation}
where $\EE_{\mu,\pi}$ denotes the expectation with respect to the randomness of the state-action pair $\{(S_h,A_h)\}_{h=1}^H$ when $S_1$ is generated from $\mu$, $A_h$ follows the policy $\pi_h(\,\cdot \cond S_h)$ and $S_{h+1}$ is generated from the transition probability $\bP_h(\,\cdot \cond S_h,A_h)$. Similarly, we define $\bP_{\mu,\pi}$ as the distribution of $\{(S_h,A_h)\}_{h=1}^H$ generated as above. 
Furthermore, we use $\EE_{\pi}$ and $\bP_\pi$ to highlight the distribution determined by the policy only when the starting state (and action) is explicitly specified.
We do not require an explicit form of $\bP_h$ and $r_h$, but a simulator with which we can query any triplet $(x,a,h) \in \bS \times \bA \times [H]$ and get a state $x'\sim \bP_h(\,\cdot \cond x,a)$ and a reward $r = r_h(x,a)$. Our goal is to find an $\epsilon$-optimal policy $\pi
^\epsilon$ such that
\begin{equation}
    J(\pi^\epsilon) \ge \sup_{\pi \in \mathcal{P}(\bA \cond \bS)} J(\pi) - \epsilon,
\end{equation}
using as fewer samples (the number of times to access the simulator) as possible. 

We define the value function $V_h^\pi:\bS \mapsto \bR$ and action-value function (Q-function) $Q_h^\pi:\bS \times \bA \mapsto \bR$ as the expected cumulative reward of the MDP starting from step $h$:
\begin{align}
    V_h^\pi(x) &= \EE_{\pi}[\sum_{h'=h}^Hr_{h'}(S_{h'},A_{h'}) \cond S_h = x],  \\
    Q_h^\pi(x,a) &=\EE_{\pi}[\sum_{h'= h}^H r_{h'}(S_{h'},A_{h'}) \cond S_h = x, A_h = a].
\end{align}
We have the following Bellman equation:
\begin{align}
    &Q_h^\pi(x,a) = (\mathcal{T}_h V_{h+1}^\pi)(x,a), \\
    &V_h^\pi(x) = \langle Q_h^\pi(x, \,\cdot\,), \pi_h(\,\cdot \cond x) \rangle_\bA,  
\end{align}
where $\langle\,\cdot \, ,\,\cdot\,\rangle_\bA$ denotes the inner product over $\bA$ and $\mathcal{T}_h f$ denotes
\begin{align}
    &(\mathcal{T}_h f)(x,a) = r_h(s,a) + \EE_{x'\sim \bP_h(\,\cdot \cond x,a)}f(x'),\; h\in[H-1], \notag \\
    &(\mathcal{T}_H f)(x,a) = r_H(s,a)
\end{align}
for any bounded measurable function $f$ on $\bS$. With our assumption on $r_h$, it is obvious to see that $Q_h^\pi$ and $V_h^\pi \in [0,H]$ for any $\pi \in \mathcal{P}(\bA \cond \bS)$ and $h \in [H]$. Since the action space and the episode length are both finite, there exists an optimal policy $\pi^{*}$ which gives the optimal value $V_h^{*}(x) = \sup_{\pi \in \mathcal{P}(\bA \cond \bS)} V_h^\pi(x)$ for any $x \in \bS$ (see, e.g. \cite[Theorem~4.3.3]{puterman2014markov}) and hence $\pi^{*}$ is also the maximizer of $J_\mu(\pi)$ on $\mathcal{P}(\bA \cond \bS)$. Moreover, we have the following Bellman optimality equation:
\begin{equation}
\begin{aligned}\label{Bellman_optimality_equation}
    &Q_h^{*}(x,a) = (\mathcal{T}_h V_{h+1}^{*})(x,a), \\
    &V_h^*(x) = \max_{a \in \bA}Q_h^{*}(x,a), 
\end{aligned}
\end{equation}
where we use the optimal value function $V_h^{*}$ and the optimal Q-function $Q_h^{*}$ to denote $V_h^{\pi^*}$ and $Q_h^{\pi^{*}}$
for short and $\pi^{*}$ can be obtained through greedy policies with respect to $\{Q_h^{*}\}_{h=1}^H$, that is 
\begin{align}
    ~\{a \in \bA, \pi_h(a\cond x) > 0\}
    \subset ~\{a \in \bA, Q_h^{*}(x,a) = \max_{a'} Q_h^{*}(x,a')\}
\end{align}
for any $x \in \bS$ and $h \in [H]$. Given the boundedness of Q-function, we also define operators $\mathcal{T}_h^{*}$
\begin{equation}
    \mathcal{T}^{*}_h f = \mathcal{T}_h \mathcal{K}_{H-h} (\max_{a \in \bA} f(\,\cdot\,,a)), 
\end{equation}
for any measurable function $f$ on $\bS \times \bA$, where $\mathcal{K}_{H-h}$ is the truncated operator defined in \eqref{truncated_operator}. If we define $Q_{H+1}^{*} = 0$, the Bellman optimality equation can be rewritten as
\begin{equation}
    Q_{h}^* = \mathcal{T}^{*}_h Q^{*}_{h+1}
\end{equation}
for any $h \in [H]$.

\subsection{Kernel Method and Reproducing Kernel Hilbert Space}
When we use the kernel method to approximate $Q_h^{*}$, the corresponding function space we consider is the reproducing kernel Hilbert space (RKHS). Since $\bA$ is finite, there are two natural choices for  formulating the function approximation problem. One is to treat $Q_h^{*}$ as a function of the state-action pair $(x,a)$ and assume that $Q_h^{*}$ is in a RKHS on $\bS \times \bA$. The other is to treat $Q_h^{*}$ as a function of $x$ given any $a \in \bA$ and assume $Q_h^{*}(\,\cdot\,,a)$ is in a RKHS on $\bS$ for any $a \in \bA$. We choose the  second formulation for generality. 
Similar results can be derived for the first formulation without any difficulty.  
Given a continuous positive definite kernel $k$ that satisfies 
\begin{enumerate}
    \item $k(x,y) = k(y,x)$, $\forall x, y \in \bS$;
    \item $\forall m \ge 1$, $x_1,\dots,x_m \in \bS$ and $a_1,\dots,a_m \in \bR$, we have:
    \begin{equation}
        \sum_{i=1}^m\sum_{j=1}^ma_ia_jk(x_i,x_j) \ge 0.
    \end{equation}
\end{enumerate}
Then, there exists a Hilbert space  $\mathcal{H}_k \subset C(\bS)$ such that
\begin{enumerate}
    \item $\forall x \in \bS$, $k(x,\,\cdot\,) \in \mathcal{H}_k$;
    \item $\forall x \in \bS$ and $f \in \mathcal{H}_k$, $f(x) = \langle f, k(x,\,\cdot\,)\rangle_{\mathcal{H}_k}$,
\end{enumerate}
and $k$ is called the reproducing kernel of $\mathcal{H}_{k}$ \cite{aronszajn1950theory} and we use $\|\cdot\|_{\mathcal{H}_k}$ to denote the norm of the Hilbert space $\mathcal{H}_k$.

We make the following assumption when we use the kernel method to make function approximation in solving the reinforcement learning problem.
\begin{assu}[RKHS]
\label{assumption_rkhs}
    There exist positive constants $K_r$, $K_p$ and probability distributions $\{\rho_{h,a}\}_{h \in [H], a\in \bA}$  for any $h \in [H]$, $a \in \bA$ and $x' \in \bS$,
    \begin{equation}
        \|r_h(\,\cdot\,,a)\|_{\mathcal{H}_k} \le K_r, \quad \|p_h(\,\cdot\,,a,x')\|_{\mathcal{H}_k} \le K_p, 
    \end{equation}
    where $p_h(x,a,x') = \frac{\rmd \bP_h(x' \cond x,a)}{\rmd \rho_{h,a}(x')}$, the Radon-Nikodym derivative between $\bP_h(\,\cdot\cond x,a)$ and $\rho_{h,a}(\,\cdot\,)$. 
    In addition, $K_x = \sup_{x \in \bS}k(x,x) < +\infty$. 
\end{assu}
\begin{rem}
This assumption can be viewed as a generalization of the assumption for linear MDP used in~\cite{bradtke1996linear,melo2007q,jin2020provably}.
When $\mathcal{H}_k$ is a finite dimensional space, let $N = \mathrm{dim}(\mathcal{H}_k)$. This assumption implies that there exists $\phi^1,\dots,\phi^N \in \mathcal{H}_k$, $r_{a,h}^1,\dots,r_{a,h}^N \in \bR$ and  signed measures $\mu_{h,a}^1,\dots,\mu_{h,a}^N$ on $\bS$, such that
\begin{align}
    &r_h(x,a) = \sum_{i=1}^N r_{a,h}^i\phi^i(x),  \notag \\
    &\bP_h(x' \cond x,a) = \sum_{i=1}^N \phi^i(x)\mu_{h,a}^i(x').
\end{align}
We recover the setting of the linear MDP. 
\end{rem}
\begin{rem}
Under the above assumption, we have $\|Q_h^\pi(\,\cdot\,,a)\|_{\mathcal{H}_k} \le K_r + HK_p$, based on the fact that
\begin{equation}
    Q_h^\pi(x,a) = r_h(x,a) + \int_{\bS} p_h(x,a,x')V_{h+1}^\pi(x')\rmd \rho_{h,a}(x')
\end{equation}
and $V_{h+1}^\pi \in [0,H]$ for any $\pi \in \mathcal{P}(\bA \cond \bS)$.
\end{rem}

\subsection{Two-layer Neural Network Models and Barron Space}
We will also consider the situation when two-layer neural networks are used for function approximation.
To this end, we consider the case when $Q_h^{*}(\,\cdot\,,a)$ is in the Barron space \cite{ma2019priori,weinan2019barron} on $\bS$ for any $a \in \bA$. To do that, we assume $\bS = \mathbb{S}^{d-1}$ in this subsection for convenience. Similar results can be obtained when $\bS$ is a general compact subset of $\bR^d$.

Given $\rho \in \mathcal{P}(\mathbb{S}^{d-1})$ and $b \in L^1(\rho)$, we can define a function on $\bS$ as
\begin{equation}\label{def_Barron_function}
    f(x) = \int_{\mathbb{S}^{d-1}}b(\omega)\sigma(\omega\cdot x) \rmd\rho(\omega),
\end{equation}
which serves as the continuous analog of a two-layer neural network
\begin{equation}
    \frac{1}{m}\sum_{i=1}^m b_j\sigma(\omega_j\cdot x)
\end{equation}
as the width $m$ goes to infinity.
 We will use the ReLU  activation function $\sigma(x) = \max \{x,0\}$ in this paper.
The Barron space\footnote{
We remark that $\mathcal{B}$ is not dense in $C(\mathbb{S}^{d-1})$ due to the lack of the bias term in each neuron~(cf. Appendix D.1 in \cite{bach2017breaking} when considering the Legendre polynomial).
One way to improve is to define $\mathcal{B}'$ by replacing $\sigma$ with $\sigma^2$ in $\mathcal{B}$. Then, by Proposition 2 or 3 in \cite{bach2017breaking}, $\mathcal{B} + \mathcal{B}'$ is dense in $C(\mathbb{S}^{d-1})$. Since considering $\mathcal{B} + \mathcal{B}'$ together does not bring any additional difficulty, we will focus on $\mathcal{B}$ for simplicity.
}
is defined as follows
\begin{align}
    \mathcal{B} = \bigg\{ f(x) = \int_{\mathbb{S}^{d-1}}b(\omega)\sigma(\omega\cdot x)\rmd\rho(\omega),~\rho \in \mathcal{P}(\mathbb{S}^{d-1})
    \text{ and } \int_{\mathbb{S}^{d-1}}|b(\omega)|\rmd \rho(\omega) < +\infty \bigg\}
\end{align}
with the norm
\begin{equation}
    \|f\|_\mathcal{B} = \inf_{b,\rho} \int_{\mathbb{S}^{d-1}}|b(\omega)|\rmd \rho(\omega)
\end{equation}
where the infimum is taken over all possible $\rho \in \mathcal{P}(\mathbb{S}^{d-1})$ and $b \in L^1(\rho)$  such that Eq.~\eqref{def_Barron_function} is satisfied. 

We have the following relationship between the RKHS and Barron space:
\begin{equation}
    \mathcal{B} =  \mathop{\cup}_{\pi \in \mathcal{P}(\mathbb{S}^{d-1})} \mathcal{H}_{k_\pi},
\end{equation}
where $k_\pi(x,y) = \EE_{\omega\sim\pi} [\sigma(\omega\cdot x)\sigma(\omega\cdot y)]$. We shall also point out that compared to the RKHS, the Barron space is much larger in high dimensions, see Example 4.3 in \cite{weinan2021kolmogorov}. We refer to \cite{weinan2019barron} for more details and properties of the Barron space.

Similar to Assumption~\ref{assumption_rkhs} (RKHS), we assume the following assumption in the Barron space setting.
\setcounter{assu}{0}
\begin{assu}[Barron]
\label{Assumption_Barron}
There exists positive constants $B_r$, $B_p$ and probability distributions $\{\rho_{h,a}\}_{h \in [H], a\in \bA}$  such that for any $h \in [H]$, $a \in \bA$ and $x' \in \bS$,
    \begin{equation}
        \|r_h(\,\cdot\,,a)\|_\mathcal{B} \le B_r,  \quad
        \|p_h(\,\cdot\,,a,x')\|_\mathcal{B} \le B_p, 
    \end{equation}
    where $p_h(x,a,x') = \frac{\rmd \bP_h(x' \cond x,a)}{\rmd \rho_{h,a}(x')}$.
\end{assu}
\begin{rem}
Similar to the RKHS setting, under the above assumption, $\|Q_h^\pi(\,\cdot\,,a)\|_\mathcal{B} \le B_r + HB_p$ for any $h \in [H]$, $a \in \bA$ and $\pi \in \mathcal{P}(\bA \cond \bS)$.
\end{rem}
\section{Fitted Q-iteration with Regularization and Main Results}
In this section, we first describe our fitted Q-iteration algorithm with regularization, given a general choice of function spaces for function approximation. Under general assumptions on the function spaces and concentratability of data distributions, we give our main theorem for the error estimation of the value function. In two subsequent subsections, we verify that both RKHS and Barron space satisfy the general assumptions thus ensure the established bound for the optimal policy.

To solve the optimal policy, one possible approach is to first compute the optimal action-value function $Q_h^{*}$ backwardly through the Bellman optimality equation \eqref{Bellman_optimality_equation} and then use the greedy policies with respect to $Q_h^{*}$. However, since we do not know an explicit form of $\mathbb{P}_h$ and $r_h$%
, the exact computation of $Q_h^{*}$ is infeasible. Therefore, we need to learn $Q_h^{*}$ using data. One observation from \eqref{Bellman_optimality_equation} is that, for any $\mu \in \mathcal{P}(\bS \times \bA)$, $Q_h^{*}$ is the minimizer of the following problem
\begin{equation}
    \min_{f \in \mathcal{F}}\,\EE_{(x,a)\sim \mu}|f(x,a) -(\mathcal{T}_h V_{h+1}^{*})(x,a)|^2,
\end{equation}
for any function class $\mathcal{F}$ such that $Q_h^{*} \in \mathcal{F}$.
With the bias-variance decomposition
\begin{align}
&\EE_{(x,a)\sim \mu, x'\sim \mathbb{P}_h(\,\cdot\cond x,a)}|f(x,a) -r_h(x,a) - V_{h+1}^{*}(x')|^2 \notag \\
=\;&\EE_{(x,a)\sim \mu, x'\sim \bP_h(\,\cdot\cond x,a)}|V_{h+1}^{*}(x') - \EE[ V_{h+1}^{*}(x') \cond x,a]|^2 + \EE_{(x,a)\sim \mu}\cond f(x,a) -( \mathcal{T}_h V_{h+1}^{*})(x,a)|^2,
\end{align}
and the observation that $f$ is not involved in the first term, 
we can deduce that $Q_h^{*}$ is the minimizer of 
\begin{equation}\label{argmin property}
     \min_{f \in \mathcal{F}}\,\EE_{(x,a)\sim \mu,x'\sim\mathbb{P}_h(\cdot|x,a)}|f(x,a) -r_h(x,a) -  V_{h+1}^{*}(x')|^2.
\end{equation}
This motivates us to consider the following algorithm: at each step $h$, we collect $n$ samples $\{(x^i,a^i)\}_{1\le i \le n}$ from a fixed distribution $\nu_h$, submit the queries $\{(x^i,a^i,h)\}_{1\le i \le n}$ to the simulator and obtain $\{r_h^i,(x^i)'\}_{1\le i \le n}$. 
We solve an empirical version of the least square problem \eqref{argmin property} for $h = H, H-1,\dots,1$ in a backward fashion. This is called fitted Q-iteration \cite{ernst2005tree,szepesvari2010algorithms,chen2019information,fan2020theoretical}.
We refer to Chapter 3 of \cite{fan2020theoretical} for more detailed discussions on the relationship between fitted Q-iteration and other reinforcement learning algorithms.

\subsection{A General Framework of Algorithms and Analysis}
Now we are ready to state our fitted Q-iteration with regularization, as detailed in Algorithm~1 below. Here $\Lambda: \mathcal{F} \mapsto [0,+\infty)$ denotes a regularization term on the approximating function.

\begin{algorithm}[ht]
\label{alg}
\caption{Fitted Q-Iteration Algorithm with Regularization}
\KwIn{MDP($\bS,\bA,\bH,\bP,r$), function classes $\mathcal{F}$, regularization term $\Lambda$, regularization constant $\lambda > 0$, sequence of sampling distributions $\{\nu_h\}_{h=1}^H$,  number of samples $n$.\\
\textbf{Initialize:} $Q_{H+1}(x,a) = 0$ for any $(x,a) \in \bS \times \bA$. }

\For{$h = H,H-1,\dots,1$}{
 
\,\,\,Sample i.i.d. $(S_h^1,A_h^1),\dots,(S_h^n,A_h^n)$ from $\nu_h $\\
Send $(S_h^1, A_h^1, h),\dots,(S_h^n,A_h^n,h)$ to the simulator
and obtain the rewards and next states for all the state-action pairs  $(r_h^1,\hat{S}_{h+1}^1), \dots,(r_h^n,\hat{S}_{h+1}^n)$\\
 Compute $y_h^i = r_h^i + \max_{a' \in \bA} Q_{h+1}(\hat{S}_{h+1}^i,a')$\\
Computing $\overline{Q}_h$ as the minimizer of the optimization problem
\begin{equation}\label{optimization_problem}
   \min_{f \in \mathcal{F}}\left\{\frac{1}{2n}\sum_{i=1}^n|y_h^i - \mathcal{K}_{H-h+1} f(S_h^i,A_h^i)|^2 + \lambda\Lambda(f)\right\} \;
\end{equation}
Set $Q_h = \mathcal{K}_{H-h+1} \overline{Q}_h$.
}
\KwOut{$\hat{\pi}$ as the greedy policies with respect to $\{Q_h\}_{h=1}^H$.}
\end{algorithm}

Compared to algorithms in \cite{chen2019information,fan2020theoretical},
we introduce a regularization term in the empirical version of the optimization problem 
\begin{align}
    \min_{f \in \mathcal{F}}\Big\{\EE_{(x,a)\sim \mu,x'\sim\mathbb{P}_h(\,\cdot \cond x,a)}|f(x,a) -r_h(x,a) -  V_{h+1}^{*}(x')|^2 + \lambda \Lambda(f)\Big\},
\end{align}
To see the benefit of the regularization term, we take $\mathcal{F}$ as a very large function class, e.g. the collection of all two-layer neural networks. The algorithms in \cite{chen2019information,fan2020theoretical} can be viewed as choosing a bounded subset of $\mathcal{F}$ as the hypothesis spaces, e.g. the two-layer neural networks with bounded weights, and solve an empirical version of the least-squares problem \eqref{argmin property}. Our algorithm uses $\mathcal{F}$ as the hypothesis space without any restriction but adds a regularization term into the optimization problem, which is closer to the practice.

To see the main ingredients in the analysis, we first state our general assumptions about $\mathcal{F}$, $\Lambda$ and $\{\nu_h\}_{h=1}^H$. These assumptions are concerned with the approximation error and estimation error in our setting. Later we will discuss two specific settings in which these assumptions are satisfied with an appropriate choice of the constants $\epsilon_f, R, M$, etc.

\setcounter{assu}{0}
\begin{assu}[General]
\label{main_assumption}
\begin{enumerate}[(i)]
    \item \label{ass11}
    There exist constants $\epsilon_f \ge 0$ and $R\ge 0$,  such that, for any $h \in [H]$ and $f\in \mathcal{F}$,  
    \begin{equation}\label{approximation_property}
        \inf_{g \in \mathcal{F}_R}\|\mathcal{T}_{h}^{*}f - g\|_{2,\nu_h} \le \epsilon_f,
    \end{equation}
    where $\mathcal{F}_r = \{f \in \mathcal{F}, \Lambda(f) \le r\}$ for any $r \in [0,+\infty)$.
    \item \label{ass12} For any integer $n \ge 1$ and $r \in (0,+\infty)$, there exists a positive constant $M\ge 1$,  such that the Rademacher complexity
    \begin{equation}
        \mathrm{Rad}_n(\mathcal{F}_r) =\frac{1}{n} \EE \sup_{f \in \mathcal{F}_r}\sum_{i=1}^n\xi_i f(x_i) \le \frac{M}{\sqrt{n}}r,
    \end{equation}
    where $\xi_1, \dots,\xi_n$ are i.i.d. random variables drawn from the Rademacher variable i.e. $\bP(\xi_i = 1) = \bP(\xi_i = -1) = \frac{1}{2}$ and $x_1,\dots,x_n$ are i.i.d. drawn from $\nu_h$ which are independent of $\xi_1,\dots,\xi_n$.
\end{enumerate}
\end{assu}
\begin{rem}\label{rem_1}
Assumption \ref{main_assumption}~\ref{ass11} is used to control the approximation error in fitted Q-iteration.
Similar assumptions are made in \cite{chen2019information,fan2020theoretical,wang2020provably}. 
See \cite{chen2019information,fan2020theoretical,wang2020provably} for more discussions on this assumption. Assumption \ref{main_assumption}~\ref{ass12} is used to control the generalization error. 
We remark that in this paper we do not make the direct assumptions on the $Q$-function in the form of Assumption \ref{main_assumption}~\ref{ass11}. Instead we only make more concrete assumptions on $\mathbb{P}$ and $r$ to ensure Assumption \ref{main_assumption}~\ref{ass11} holds (see Propositions~\ref{Proposition_RKHS} and \ref{Proposition_Barron}).
Rademacher complexity in Assumption \ref{main_assumption}~\ref{ass12} is widely used to control the generalization error in machine learning \cite{bartlett2002rademacher,shalev2014understanding,ma2019priori}. We refer to \cite{bartlett2002rademacher,shalev2014understanding} for more discussions on Rademacher complexity.
\end{rem}

\begin{assu}
\label{assumption_concentration}
For any policy $\pi \in \mathcal{P}(\bA \cond \bS)$, we consider the MDP $(S_h,A_h)_{h=1}^H$ generated from $(\nu_1,\bP,\pi)$, i.e. $(S_1, A_1) \sim \nu_1$ and $S_{h+1} \sim \bP_{h}(\,\cdot \cond S_h,A_h)$ and $A_{h+1} \sim \pi_{h+1}(\, \cdot \cond S_{h+1})$ for $h \in [H-1]$. We use $\bP_h^\pi \nu_1$ to denote the distribution of $(S_{h},A_h)$. Then we assume the following concentration coefficients $\{\kappa_h\}_{h=2}^H$ are all finite:
    \begin{equation}
        \kappa_h \coloneqq \sup_{\pi \in  \mathcal{P}(\bA \cond \bS,H)}\left[\EE_{\nu_h}\left| \frac{\rmd \bP_h^\pi \nu_1}{\rmd \nu_h}\right|^2\right]^{\frac{1}{2}} < +\infty.
    \end{equation}
    We assume without loss of generality that $\kappa_1 = 1$. In addition, let
    \begin{equation}
        \kappa = H^{-2}\sum_{h=1}^H h\kappa_h.
    \end{equation}
\end{assu}

\begin{rem}\label{rem_2}    
    Assumption \ref{assumption_concentration} measures the maximum difference between $\nu_h$ and all possible distributions of $(S_h,A_h)$ with admissible policies. A sufficient condition for Assumption \ref{assumption_concentration} is that there exists a constant $C> 0$ and $\{\mu_{h}\}_{h=2}^H \in \mathcal{P}(\bS)$ such that $\bP_h(\,\cdot \cond x,a) \le C\mu_{h+1}(\,\cdot\,)$ for any $(x,a,h)\in \bS \times \bA \times [H-1]$. In this case, 
    taking $\nu_h = \mu_h \times \mu_\bA$, where $\mu_\bA$ is the uniform distribution on $\bA$, we have: 
    \begin{align}
        & (\bP_h^\pi\nu_1)(E,a) \notag \\
        =~&  \int_{\bS \times \bA}\bP_{h-1}(E \cond x',a')\pi_h(a \cond x)\rmd \bP_{h-1}^\pi  \nu_1(x',a') \notag \\
        \le~& C|\bA| \nu_{h}(E,a),\notag
    \end{align}
    for any $h \ge 2$, $a \in \bA$ and measurable set $E \subset \bS$, which means that $\frac{\rmd \bP_h^\pi\nu_1}{\rmd \nu_h} \le C|\bA|$. 
    We can then conclude that $\kappa_h \le C|\bA|$ and $\kappa \le C|\bA|$. Similar assumptions are made in \cite{munos2008finite,farahmand2010error,scherrer2015approximate,farahmand2016regularized,lazaric2016analysis,chen2019information,fan2020theoretical}. 
    We refer to \cite{chen2019information} for the necessity of this type of assumptions in batch reinforcement learning and more discussions. 
    We shall note that Assumption \ref{assumption_concentration} implies $|\bA| < \infty$. To see this, we consider a policy $\pi$ that always takes action $a^*$ at step $h$ regardless of the state.  By the Cauchy–Schwarz inequality and Assumption \ref{assumption_concentration}, we have
    \begin{equation*}
        \nu_h(\bS\times\{a^*\})\kappa_h^2 \ge \EE_{\nu_h}[\mathrm{1}_{a = a^*}] \EE_{\nu_h}\left| \frac{\rmd \bP_h^\pi \nu_1}{\rmd \nu_h}\right|^2 \ge \left|\EE_{\nu_h}\frac{\rmd \bP_h^\pi \nu_1}{\rmd \nu_h} \mathrm{1}_{a = a^*}\right|^2 = 1.
    \end{equation*}
    Therefore, $|\bA| \leq \max_{2\le h\leq H}\kappa_h^2 < \infty$.
    Similarly, Assumption \ref{assumption_concentration} also rules out non-trivial deterministic MDPs. In deterministic MDPs, $\bP_h^\pi \nu_1$ must be a Dirac delta function. With an argument similar to the above, we can prove that in the deterministic MDPs, $\kappa_h$ can never be finite, unless $\{\bP_h^\pi \nu_1\}_{\pi \in \mathcal{P}(\bA \cond \bS)}$ is finite or the set of all possible visiting states is finite.
\end{rem}
\begin{thm}\label{main_theorem}
Under Assumptions~\ref{main_assumption} (General) and~\ref{assumption_concentration}, for any $\delta \in (0,1)$, if $\lambda \ge \frac{2MH}{\sqrt{n}}$ for any $h \in [H]$, then with probability $1-\delta$, we have
\begin{align}
     &\|V_1^{*}- V_1^{\hat{\pi}}\|_{1,\nu_1} \notag \\
     \le~&  2\kappa H^2\Bigg\{\epsilon_f^2 + 2[\lambda + \frac{2MH}{\sqrt{n}}](R+1) + 4H^2\Big[2\sqrt{\frac{\ln(4H/\delta)}{n}} + \frac{2}{\sqrt{n}} +  \sqrt{\frac{\ln(8nH(H+R)/\delta)}{n}}\Big]\Bigg\}^\frac{1}{2},
\end{align}
where $\hat{\pi}$ is the output of Algorithm 1.
In addition, let $\nu$ be the marginal distribution of $\nu_1$ on $\bS$, we have
\begin{align}
    & \max_{\pi \in \mathcal{P}(\bA \cond \bS)}J_\nu(\pi) -  J_\nu(\hat{\pi})\notag \\
    \le~&  2\kappa H^2\Bigg\{\epsilon_f^2 + 2[\lambda + \frac{2MH}{\sqrt{n}}](R+1) + 4H^2\Big[2\sqrt{\frac{\ln(4H/\delta)}{n}} + \frac{2}{\sqrt{n}} +  \sqrt{\frac{\ln(8nH(H+R)/\delta)}{n}}\Big]\Bigg\}^\frac{1}{2}. 
\end{align}
\end{thm}
\begin{rem}
When $\epsilon_f = 0$ and take $\lambda = \frac{2MH}{\sqrt{n}}$, Theorem \ref{main_theorem} implies that Algorithm 1 can obtain an $\tilde{O}(\kappa H^2 [H+(MHR)^\frac{1}{2}] n^{-\frac{1}{4}})$-optimal policy with $Hn$ samples with high probability.  
\end{rem}

\subsection{RKHS Setting}
For the RKHS setting, we choose
\begin{align}
    &\mathcal{F} = \{ f \in C(\bS \times \bA), f(\,\cdot\,,a) \in \mathcal{H}_k \text{ for any } a \in \bA\}, \\
    &\Lambda(f) = \max_{a \in \bA}\|f(\,\cdot\,,a)\|_{\mathcal{H}_k}.
\end{align}

\begin{prop}\label{Proposition_RKHS}
Under Assumption \ref{assumption_rkhs} (RKHS), 
 Assumption \ref{main_assumption} (General) is satisfied with $\epsilon_f = 0$, $R = K_r + HK_p$ and $M = \sqrt{|\bA| K_x}$.
\end{prop}
\begin{rem}
 From the above proposition and Theorem \ref{main_theorem}, %
We conclude that under Assumption \ref{assumption_concentration}, $\hat{\pi}$ in Algorithm 1 is an $\tilde{O}(\kappa H^3|\bA|^\frac{1}{4}n^{-\frac{1}{4}})$-optimal policy with $Hn$ samples in high probability.
\end{rem}
\begin{rem}\label{rem_7}
Assumption \ref{assumption_rkhs}~(RKHS) can be relaxed as follows: There exist positive constants $\epsilon_r$ and $\epsilon_p$ such that
\begin{align}
    &\inf_{\|g\|_{\mathcal{H}_k} \le K_r}\|g - r_h(\,\cdot\,,a)\|_{2,\nu_{h,a}} \le \epsilon_r, \\  &\inf_{\|g\|_{\mathcal{H}_k} \le K_p}\|g - p_h(\,\cdot\,,a,x')\|_{2,\nu_{h,a}} \le \epsilon_p
\end{align}
for any $h \in [H]$, $a \in \bA$ and $x' \in \bS$, where $\nu_{h,a}$ is the conditional probability of $\nu_h$ on $\bS$ given $a$. In this case, $\epsilon_f = \epsilon_r + H\epsilon_p$. Also, we can replace $K_x$ by $\sup_{h \in [H]}\int_{\bS}k(x,x)\rmd \nu_h(x,a)$.
\end{rem}
\begin{rem}
For any $f\in \mathcal{H}_k$,
\begin{equation}
    |f(x)| = |\langle f,k(x,\,\cdot\,)\rangle_{\mathcal{H}_k}| \le \sqrt{K_x}\|f\|_{\mathcal{H}_k}.
\end{equation}
Therefore, if
$\rho_h = \frac{1}{|\bA|}\sum_{a \in \bA} \rho_{h,a}$ where $\rho_{h,a}$ is stated in Assumption~\ref{assumption_rkhs} (RKHS), we can obtain that $\bP_h(\,\cdot \cond x,a) \le \sqrt{K_x}|\bA|K_p \rho_h(\,\cdot\,)$ for any $(x,a,h) \in \bS \times \bA \times [H]$. Choosing $\nu_h = \rho_{h-1} \times \mu_\bA$, Assumption \ref{assumption_concentration} is satisfied with $\kappa \le \sqrt{K_x} |\bA|^2K_p $ (See Remark \ref{rem_2}). In practice, it requires some prior knowledge about the MDP to obtain $\rho_h$.
\end{rem}

Since $\mathcal{F}$ is an infinite dimensional space, it is hard to perform the optimization problem \eqref{optimization_problem} directly. However, we can prove that if we replace $\mathcal{F}$ with
\begin{align}\label{Def_small_function}
    \hat{\mathcal{F}}_h = \{&f \in C(\bS \times \bA), \exists\;b_i \in \bR, 1\le i \le n ,\\ 
    &\text{such that } f(x,a) =\sum_{1 \le i \le n, A_h^i = a} b_i k(x,S_h^i)\},
\end{align}
the same result can be obtained (see Section \ref{Appendix_proposition_2}). Moreover, 
we can remove the truncation operator $\mathcal{K}_{H-h+1}$ in the optimization problem and hence obtain a convex optimization problem with only $n$ parameters. 

\subsection{Barron Space Setting}
For the Barron space setting, we choose
\begin{align}
    &\mathcal{F} = \Big\{f(x,a) = \frac{1}{m}\sum_{i=1}^m b_{i,a}\sigma(\omega_{i,a}\cdot x),~b_{i,a} \in \bR \text{ and }\omega_{i,a} \in \bR^d \text{ for any }a \in \bA \text{~and~} i \in [m]\Big\}, \notag \\
    &\Lambda(f) = \max_{a \in \bA}\frac{1}{m}\sum_{i=1}^m|b_{i,a}|\|\omega_{i,a}\| \text{  when }f(x,a) = \frac{1}{m}\sum_{i=1}^m b_{i,a}\sigma(\omega_{i,a}\cdot x),
\end{align}
for a fixed $m \in \mathbb{N}^{+}$.
\begin{prop}\label{Proposition_Barron}
Under Assumption~\ref{Assumption_Barron} (Barron), Assumption~\ref{main_assumption} (General) is satisfied with $R = 2(B_r +HB_p)$, $\epsilon_f = Rm^{-\frac{1}{2}}$ and $M = 2|\bA|^{\frac{1}{2}}$.
\end{prop}
\begin{rem}
From the above proposition and Theorem \ref{main_theorem},  %
we conclude that under Assumption \ref{assumption_concentration}, $\hat{\pi}$ in Algorithm 1 is an $\tilde{O}(\kappa H^3[|\bA|^\frac{1}{4}n^{-\frac{1}{4}}) + m^{-\frac{1}{2}}])$-optimal policy with $Hn$ samples in high probability.
Hence, if $m = O(n^{\frac{1}{2}}|\bA|^{-\frac{1}{2}})$, we can recover the convergence rate $\tilde{O}(\kappa H^3|\bA|^{\frac{1}{4}}n^{-\frac{1}{4}})$.
\end{rem}
\begin{rem}\label{rem_10}
Assumption \ref{assumption_rkhs} (Barron) can also be relaxed as follows: There exists positive constants $\epsilon_r$ and $\epsilon_p$ such that
\begin{align}
    &\inf_{\|g\|_\mathcal{B} \le B_r}\|g - r_h(\,\cdot\,,a)\|_{2,\nu_{h,a}} \le \epsilon_r,  \notag \\
    &\inf_{\|g\|_\mathcal{B} \le B_p}\|g - p_h(\,\cdot\,,a,x')\|_{2,\nu_{h,a}} \le \epsilon_p
\end{align}
for any $h \in [H]$, $a \in \bA$ and $x' \in \bS$. In this case, $\epsilon_f = \epsilon_r + H\epsilon_p + Rm^{-\frac{1}{2}}$.
\end{rem}
Moreover, by Assumption~\ref{Assumption_Barron} (Barron), $\bP_h(\,\cdot\,\cond x,a) \le B_p|\bA| \rho_h(\,\cdot\,)$ for any $(x,a,h) \in \mathbb{S}^{d-1} \times \bA \times [H]$ where $\rho_h = \frac{1}{|\bA|}\sum_{a \in \bA} \rho_{h,a}$. Therefore, we can make Assumption \ref{assumption_concentration} true by choosing $\nu_h = \rho_{h-1}\times \mu_\bA$ with $\kappa \le |\bA|^2B_p$ (See Remark \ref{rem_2}). Again, some prior knowledge about the MDP is required to obtain $\rho_h$. An alternative approach is to consider
\begin{equation}
    \tilde{\rho}_{h,a}(\,\cdot\,) =  \int_{\mathbb{S}^{d-1}} \bP_h(\,\cdot\cond x,a) \rmd \pi(x),
\end{equation}
where $\pi$ is the uniform distribution on $\mathbb{S}^{d-1}$. Assuming that there exist signed measures $\{\mu_{h,a,x'}\}_{h \in [H],a \in \bA,x' \in \mathbb{S}^{d-1}}$  and a constant $\tau > 0$ such that
\begin{equation}
   p_h(x,a,x') = \int_{\mathbb{S}^{d-1}}\sigma(\omega \cdot x)\rmd \mu_{h,a,x'}(\omega),
\end{equation}
and 
\begin{equation}
    \mu_{h,a,x'}(\mathbb{S}^{d-1})\ge \tau
\end{equation}
 for any $(h,a,x') \in [H]\times \bA \times \mathbb{S}^{d-1}$, 
then
\begin{align}
   \frac{\rmd\tilde{\rho}_{h,a}}{\rmd \rho_{h,a}}(x') &=  \int_{\mathbb{S}^{d-1}\times \mathbb{S}^{d-1}} \sigma(\omega \cdot x) \rmd \mu_{h,a,x'}(\omega) \rmd\pi(x) \notag \\
   &=  C_d\mu_{h,a,x'}(\mathbb{S}^{d-1}) \ge C_d \tau,
\end{align}
where $C_d = \int_{\bS^{d-1}} \sigma(\omega \cdot x)\rmd \pi(x) \ge \frac{c}{\sqrt{d}}$ for a universal constant $c > 0$. Then, $\bP_h(\,\cdot\cond x,a) \le \frac{B_p}{C_d\tau}|\bA| \tilde{\rho}_h(\,\cdot\,)$ where $\tilde{\rho}_h = \frac{1}{|\bA|}\sum_{a \in \bA} \tilde{\rho}_{h,a}$. Therefore, if we  choose $\nu_h = \tilde{\rho}_{h-1} \times \mu_\bA$,  Assumption \ref{assumption_concentration} is satisfied with $\kappa \le |\bA|^2 \frac{B_p}{C_d \tau} = O(\sqrt{d}|\bA|^2 B_p)$. Compared to $\rho_h$, $\tilde{\rho}_h$ has the benefit that it can be easily sampled even without any prior knowledge about the MDP.
\section{Proofs of Main Theorems and Supporting Propositions}
We recall some notations for readers convenience.
$\mathcal{T}_h f$ denotes
\begin{align}
    &(\mathcal{T}_h f)(x,a) = r_h(s,a) + \EE_{x'\sim \bP_h(\,\cdot \cond x,a)}f(x'),\; h\in[H-1], \notag \\
    &(\mathcal{T}_H f)(x,a) = r_H(s,a)
\end{align}
for any bounded measurable function $f$ on $\bS$.
Given $m > 0$, $\mathcal{K}_m$ is a truncation operator defined as 
$\mathcal{K}_mf(x) = \min\{ \max\{f(x),0\},m\}$.
We also define an operator $\mathcal{T}_h^{*}$
\begin{equation}
    \mathcal{T}^{*}_h f = \mathcal{T}_h \mathcal{K}_{H-h} (\max_{a \in \bA} f(\,\cdot\,,a)), 
\end{equation}
for any measurable function $f$ on $\bS \times \bA$. 
By the boundedness of the reward function, we have the Bellman optimality equation 
\begin{equation}
    Q_{h}^* = \mathcal{T}^{*}_h Q^{*}_{h+1}
\end{equation}
for any $h \in [H]$ ($Q_{H+1}^{*} = 0$).
\subsection{Proof of Theorem 1}
We recall that $\overline{Q}_h$ denotes the minimizer of the optimization problem \eqref{optimization_problem} and $Q_h=\mathcal{K}_{H-t+1}\overline{Q}_h$ denotes the approximating Q-function in Algorithm 1.

Our first goal is to estimate the one-step error
\begin{equation}
    \|\mathcal{T}_{h}^{*}Q_{h+1} - Q_{h}\|_{2,\nu_h}, \text{ for any } h \in [H].
\end{equation}

Let
\begin{align}
    L_n(f) &= \frac{1}{2n}\sum_{i=1}^n|r_h(S_h^i,A_h^i) + V_{h+1}(\hat{S}^i_{h+1}) - \mathcal{K}_{H-h+1}f(S_h^i,A_h^i)|^2, \notag \\
    L(f) &= \frac{1}{2}\EE_{(x,a) \sim \nu_h,x'\sim \bP_h(\,\cdot \cond x,a)}|r_h(x,a) + V_{h+1}(x') - \mathcal{K}_{H-h+1}f(x,a)|^2,
\end{align}
where $V_h(x) = \max_{a \in \bA} Q_{h}(x,a)$ for $h \in [H]$ and $V_{h+1} = 0$.
By definition $\EE L_n(f) = L(f)$.

Since $r_h(x,a) + V_{h+1}(x')  \in [0,H]$, 
we know that $\frac{1}{2}|r_h(S_h^i,A_h^i) + V_{h+1}(\hat{S}^i_{h+1}) - \mathcal{K}_{H-h+1} f|^2$ is $H$-Lipschitz with respect to $f$ and bounded by $H^2$. Then, following Theorem 26.5 and Lemma 26.9 in \cite{shalev2014understanding} and Assumption \ref{main_assumption} (General)~\ref{ass12}, for any $r \in (0,+\infty)$ and $\delta \in (0,1)$, with probability at least $1-\delta$, we have
\begin{equation}\label{Generalization_class}
    \sup_{f \in \mathcal{F}_r}|L(f) - L_n(f)| \le \frac{2MH}{\sqrt{n}}r + H^2\sqrt{\frac{2\ln(2/\delta)}{n}}.
\end{equation}
Noticing that $\mathcal{F} = \cup_{l\in \mathbb{N}^{+}} \mathcal{F}_l$, we choose $l=1,2,\dots$ and $\delta_l = \frac{\delta}{2l^2}$ to get, with probability at least $1-\delta_l$,
\begin{equation}
      \sup_{f \in \mathcal{F}_l}|L(f) - L_n(f)| \le \frac{2MH}{\sqrt{n}}l + H^2\sqrt{\frac{2\ln(4l^2/\delta)}{n}}
\end{equation}
holds true. Hence, with probability at least
\begin{equation}
    1 - \sum_{l=1}^{\infty}\delta_l = 1-\frac{\delta}{2}\sum_{l=1}^{\infty}\frac{1}{l^2} \ge 1 - \delta,
\end{equation}
we have
\begin{equation}
      \exists\;l\in \mathbb{N}^{+} \text{ such that }  \sup_{f \in \mathcal{F}_l}|L(f) - L_n(f)| \le \frac{2MH}{\sqrt{n}}l + H^2\sqrt{\frac{2\ln(4l^2/\delta)}{n}}.
\end{equation}
For any $f \in \mathcal{F}$, there exists $l_0 \in \mathbb{N}^+$ such that $l_0 -1 < \Lambda(f) \le l_0$. Therefore we have with probability at least $1-\delta$, for any $f \in \mathcal{F}$,
\begin{equation}\label{Generalization_Regularization}
    |L(f) - L_n(f)| \le \frac{2MH}{\sqrt{n}}l_0 +H^2\sqrt{\frac{2\ln(4l_0^2/\delta)}{n}} \le  \frac{2MH}{\sqrt{n}}(\Lambda(f) + 1) +H^2\sqrt{\frac{2\ln(4(\Lambda(f) + 1)^2/\delta)}{n}}.
\end{equation}

Using inequality \eqref{Generalization_class}, we have with probability at least $1-\frac{\delta}{2}$,
\begin{equation}
    \min_{f \in \mathcal{F}_R} L_n(f) \le \min_{f \in \mathcal{F}_R} L(f) + \sup_{f \in \mathcal{F}_R} [L_n(f) - L(f)] 
    \le \min_{f \in \mathcal{F}_R} L(f) + \frac{2MH}{\sqrt{n}}R + H^2\sqrt{\frac{2\ln(4/\delta)}{n}}.
\end{equation}
Hence, we have with probability at least $1-\frac{\delta}{2}$,
\begin{equation}\label{Q-Estimation}
   L_n(\overline{Q}_h) + \lambda\Lambda(\overline{Q}_h) =  \min_{f \in \mathcal{F}} [L_n(f) + \lambda\Lambda(f)]\le \min_{f \in \mathcal{F}_R}L(f) +\lambda R+ \frac{2MH}{\sqrt{n}}R + H^2\sqrt{\frac{2\ln(4/\delta)}{n}}.
\end{equation}
Combing the last inequality and inequality \eqref{Generalization_Regularization} (replacing $\delta$ with $\frac{\delta}{2}$), we have
\begin{align}
    L(\overline{Q}_h) &\le   \min_{f \in \mathcal{F}_R}L(f) +\lambda R+ \frac{2MH}{\sqrt{n}}R + H^2\sqrt{\frac{2\ln(4/\delta)}{n}} -\lambda \Lambda(\overline{Q}_{h})+ \frac{2 MH}{\sqrt{n}}(\Lambda(\overline{Q}_{h}) +1)\notag \\
    &\quad + H^2\sqrt{\frac{2\ln(8(\Lambda(\overline{Q}_{h}) + 1)^2/\delta)}{n}}
\end{align}
holds true with probability at least $1-\delta$.
When $\lambda \ge \frac{2MH}{\sqrt{n}}$, with the same high probability, we have
\begin{equation}
\label{L_overlineQ_estimate}
     L(\overline{Q}_{h}) \le   \min_{f \in \mathcal{F}_R}L(f) +\lambda R+ \frac{2MH}{\sqrt{n}}R + H^2\sqrt{\frac{2\ln(4/\delta)}{n}} + \frac{2 MH}{\sqrt{n}}+ H^2\sqrt{\frac{2\ln(8(\Lambda(\overline{Q}_{h}) + 1)^2/\delta)}{n}}.
\end{equation}

Now we consider the bias-variance decomposition
\begin{align}
\label{bias-var_decomposion}
    &\EE_{(x,a) \sim \nu_h,x'\sim \bP_h(\,\cdot \cond x,a)}|r_h(x,a) + V_{h+1}(x') - f(x,a)|^2 \notag \\
    = ~&\EE_{(x,a) \sim \nu_h}|(\mathcal{T}_{h} V_{h+1})(x, a) - f(x, a)|^2 \notag \\ &+ \EE_{(x,a) \sim \nu_h,x'\sim \bP_h(\, \cdot \cond x,a)}| \EE[V_{h+1}(x') \cond x,a] - V_{h+1}(x')|^2
\end{align}
Letting $f=\overline{Q}_h$ and rearranging terms, we obtain
\begin{align}
\label{decomposition_overlineQ}
    &2L(\overline{Q}_{h}) -  \EE_{(x,a) \sim \nu_h,x'\sim \bP_h(\, \cdot \cond x,a)}| \EE[V_{h+1}(x') \cond x,a] - V_{h+1}(x')|^2  \notag \\
    =~ &  \|\mathcal{T}_{h} V_{h+1} - \overline{Q}_{h}\|_{2,\nu_h}^2 
    =  \|\mathcal{T}^{*}_{h} Q_{h+1} - \overline{Q}_{h}\|_{2,\nu_h}^2 
    \geq \|\mathcal{T}^{*}_{h} Q_{h+1} - Q_{h}\|_{2,\nu_h}^2,
\end{align}
where the last inequality is true because $\mathcal{T}^{*}_{h} Q_{h+1} \in [0, H-h+1]$ so that truncation can only reduce the squares.

By the fact $\overline{Q}_{h+1} \in \mathcal{F}$ (let $\overline{Q}_{H+1}$ be any function in $\mathcal{F}$) and Assumption~\ref{main_assumption} (General) \ref{ass11}, we know that
\begin{equation}
    \inf_{f \in \mathcal{F}_R}\|\mathcal{T}_h V_{h+1} - f\|_{2,\nu_h} =  \inf_{g \in \mathcal{F}_R}\|\mathcal{T}^{*}_h \overline{Q}_{h+1} - f \|_{2,\nu_h} \le \epsilon_f.
\end{equation}
Then taking the infimum on both sides of \eqref{bias-var_decomposion} gives us
\begin{equation}
\label{decomposition_min}
     2\min_{f \in \mathcal{F}_R}L(f) \le \epsilon_f^2 + \EE_{(x,a) \sim \nu_h,x'\sim \bP_h(\,\cdot \cond x,a)}| \EE[V_{h+1}(x')|x,a] - V_{h+1}(x')|^2.
\end{equation}
Combining \eqref{L_overlineQ_estimate}\eqref{decomposition_overlineQ}\eqref{decomposition_min}, we have with probability at least $1-\delta$
\begin{equation}\label{1_estimation_1}
    \|\mathcal{T}_{h}^{*}Q_{h+1} - Q_{h}\|_{2,\nu_h}^2 \le \epsilon_f^2 + 2[\lambda + \frac{2MH}{\sqrt{n}}](R+1) + 2H^2\Big[\sqrt{\frac{2\ln(4/\delta)}{n}} + \sqrt{\frac{2\ln(8(\Lambda(\overline{Q}_{h}) + 1)^2/\delta)}{n}}\Big].
\end{equation}
From inequality \eqref{Q-Estimation}, we know with the same high probability $1-\delta$,
\begin{align}
    \Lambda(\overline{Q}_{h}) &\le R + \frac{1}{\lambda}[\min_{f \in \mathcal{F}_R} L(f) + \frac{2MH}{\sqrt{n}}R + H^2\sqrt{\frac{2\ln(4/\delta)}{n}}]\notag\\
    & \le R + \frac{1}{\lambda}[H^2 + \frac{2MH}{\sqrt{n}}R + H^2\sqrt{\frac{2\ln(4/\delta)}{n}}]\notag\\
    &\le R + H[\sqrt{\ln(4/\delta)} +1] + \frac{H}{2\sqrt{n}}.
\end{align}
Therefore, with the same high probability, 
\begin{align}
    \sqrt{\frac{2\ln(8(\Lambda(\overline{Q}_{h}) + 1)^2/\delta)}{n}} &\le \sqrt{\frac{2\ln(8n(H+R)^2/\delta)}{n}} + \sqrt{\frac{4\ln ([(H+R)^2n]^{-\frac{1}{2}}\Lambda(\overline{Q}_{h}) + 1)}{n}} \notag \\
    &\le \sqrt{\frac{4\ln(8n(H+R)/\delta)}{n}} + \frac{2}{n}\sqrt{\frac{\Lambda(\overline{Q}_{h})}{H+R}} \notag\\
    &\le \sqrt{\frac{4\ln(8n(H+R)/\delta)}{n}} + \frac{2}{\sqrt{n}}[2 + \sqrt{\ln(4/\delta)}].
\end{align}
Combining this estimate and inequality \eqref{1_estimation_1}, we have with probability at least $1-\delta$,
\begin{equation}
     \|\mathcal{T}_{h}^{*}Q_{h+1} - Q_{h}\|_{2,\nu_h}^2 \le \epsilon_f^2 + 2[\lambda + \frac{2MH}{\sqrt{n}}](R+1) + 4H^2\Big[2\sqrt{\frac{\ln(4/\delta)}{n}} + \frac{2}{\sqrt{n}} +  \sqrt{\frac{\ln(8n(H+R)/\delta)}{n}}\Big].
\end{equation}
By taking the union with all $h \in [H]$, we conclude
\begin{align}\label{one_step_error}
     \|\mathcal{T}_{h}^{*}Q_{h+1} - Q_{h}\|_{2,\nu_h}^2 &\le \epsilon_f^2 + 2[\lambda + \frac{2MH}{\sqrt{n}}](R+1) \notag\\ & \quad+ 4H^2\Big[2\sqrt{\frac{\ln(4H/\delta)}{n}} + \frac{2}{\sqrt{n}} +  \sqrt{\frac{\ln(8nH(H+R)/\delta)}{n}}\Big]
\end{align}
holds for all $h \in [H]$ with probability at least $1-\delta$.

The next step is to deal with error propagation.
Let $\epsilon_h(x,a) = (\mathcal{T}_h^{*} Q_{h+1} - Q_h)(x,a)$. 
Since $Q_{h+1} \in [0,H-h]$, we have
\begin{equation}
    \epsilon_h(x,a)=  r_h(x,a) +  \EE_{x'\sim \bP_h(\, \cdot \cond x,a)}[\max_{a'\in\bA}Q_{h+1}(x',a')]- Q_h(x,a),
\end{equation}
and furthermore
\begin{equation}
    Q_h^{*}(x,a) - Q_h(x,a) = -\epsilon_h(x,a) + \EE_{x'\sim \bP_h(\, \cdot \cond x,a)}[\max_{a'} Q_{h+1}^{*}(x',a') - \max_{a'}Q_{h+1}(x',a')],
\end{equation}
where $Q_h^{*}$ denotes the optimal Q-function.
Hence, for any policy $\pi \in \mathcal{P}(\bA \cond \bS,H)$,
\begin{align}
    \|Q_{h}^{*}- Q_h\|_{1,\bP_h^\pi\nu_1} &\le \|\epsilon_h\|_{1,\bP_h^\pi\nu_1} + \|\EE_{x' \sim \bP_h(\, \cdot \cond x,a)}[\max_{a'} Q_{h+1}^{*}(x',a') - \max_{a'}Q_{h+1}(x',a')]\|_{1,\bP_h^\pi\nu_1}\notag \\
    &\le \|\epsilon_h\|_{1,\bP_h^\pi\nu_1} +\|\max_{a'} Q_{h+1}^{*}(x,a') - \max_{a'}Q_{h+1}(x,a')\|_{1,\bP_{h+1}^\pi\nu_1}\notag\\
    &\le  \|\epsilon_h\|_{1,\bP_h^\pi\nu_1} +\|\max_{a'} |Q_{h+1}^{*}(x,a') - Q_{h+1}(x,a')|\|_{1,\bP_{h+1}^\pi\nu_1}\notag\\
    &\le \|\epsilon_h\|_{1,\bP_h^\pi\nu_1} +\| Q_{h+1}^{*}(x,a) - Q_{h+1}(x,a)]\|_{1,\bP_{h+1}^{\tilde\pi}\nu_1},
\end{align}
where $\tilde{\pi}_{h'} = \pi_{h'}$ for $h' \neq h+1$ and $\tilde{\pi}_{h+1}(a \cond x) = \argmax_{a \in \bA} |Q_{h+1}^{*} - Q_{h+1}|(x,a)$. 
By Assumption~\ref{assumption_concentration}, have
\begin{equation}
  \|\epsilon_h\|_{1,\bP_h^\pi\nu_1} = \int_{\bS \times \bA}|\epsilon_h(x,a)| \rmd \bP_h^\pi \nu_1(x,a) \le \kappa_h \|\epsilon_h\|_{2,\nu_h}.
\end{equation}
Combining the above two inequalities, we have that for any $h \in [H]$ (letting $Q_{H+1}^* = Q_{H+1} = 0$)
\begin{equation}
    \sup_{\pi \in \mathcal{P}(\bA \cond \bS,H)}\|Q_{h}^{*}- Q_h\|_{1,\bP_h^\pi\nu_1}\le  \kappa_h\|\epsilon_h\|_{2,\nu_h} + \sup_{\pi \in \mathcal{P}(\bA \cond \bS,H)} \|Q_{h+1}^{*}- Q_{h+1}\|_{1,\bP_{h+1}^\pi\nu_1}.
\end{equation}
Recursively applying the above inequality from $h=H, H-1, \dots$, we obtain
\begin{equation}
    \sum_{h=1}^H\sup_{\pi \in \mathcal{P}(\bA \cond \bS)}\|Q_h^{*} - Q_h\|_{1,\bP_h^\pi\nu_1}\le \sum_{h=1}^H h\kappa_h\|\epsilon_h\|_{2,\nu_h}.
\end{equation}

To finally estimate
\begin{equation}
    \|V_1^{*} - V_1^{\hat{\pi}}\|_{1,\nu_1},
\end{equation}
the difference between the value functions with respect to $\pi^{*}$ and $\hat{\pi}$,
we need the classical performance difference lemma~\cite{kakade2003sample}. It is a fundamental tool in the convergence analysis of reinforcement learning and has many forms. We recall the following result \cite[Lemma~3.2]{cai2019provably} that we use below, in which $\langle\,\cdot \, ,\,\cdot\,\rangle_\bA$ denotes the inner product over $\bA$.

\begin{lem}[Performance Difference Lemma]
For any two policies $\pi$ and $\pi' \in \mathcal{P}(\bA \cond \bS)$ and any $s \in \bS$, we have
\begin{equation}
    V_1^\pi(s) - V_1^{\pi'}(s) = \EE_{\pi'}[\sum_{h=1}^H \langle Q_h^{\pi}(S_h,\,\cdot\,), \pi_h(\,\cdot \cond S_h) - \pi'_h(\,\cdot \cond S_h)\rangle_\bA \cond S_1 = s],
\end{equation}
where $\langle \,\cdot\,,\,\cdot\, \rangle_\bA$ denotes the inner product on $\bA$.
\end{lem}

We take $\pi = \pi^{*}$ and $\pi' = \hat{\pi}$ in the above lemma to obtain
\begin{align}
    0 &\le V_1^{*}(s) - V_1^{\hat{\pi}}(s) = \EE_{\hat{\pi}}[\sum_{h=1}^H\langle Q_h^*(S_h,\,\cdot\,),\pi_h^{*}(\,\cdot \cond S_h) - \hat{\pi}_h(\, \cdot \cond S_h)\rangle_\bA \cond S_1 = s] \notag \\
    &=\EE_{\hat{\pi}}[\sum_{h=1}^H\langle Q_h^*(S_h,\, \cdot \,) - Q_h(S_h, \,\cdot\,),\pi_h^{*}(\,\cdot \cond S_h) \rangle_\bA + \langle Q_h(S_h, \,\cdot\,),\pi_h^{*}(\,\cdot \cond S_h) -\hat{\pi}_h(\,\cdot \cond S_h)\rangle_\bA \notag\\
    &\quad\quad\quad\quad\;\; + \langle Q_h(S_h,\,\cdot\,) - Q_h^{*}(S_h, \,\cdot\,),\hat{\pi}_h(\,\cdot \cond S_h) \rangle_\bA  \cond S_1 = s].
\end{align}
Noticing that $\langle Q_h(s, \, \cdot \,),\pi_h^{*}(\,\cdot \cond s) - \hat{\pi}_h(\, \cdot \cond S_h)\rangle_\bA \le 0$ since $\hat{\pi}$ is the greedy policy with respect to $\{Q_h\}_{h=1}^H$, we have
\begin{equation}
    \|V_1^{*}- V_1^{\hat{\pi}}\|_{1,\nu_1} \le 2 \sum_{h=1}^H\sup_{\pi \in \mathcal{P}(\bA \cond \bS)}\|Q_h^{*} - Q_h\|_{1,\bP_h^\pi\nu_1}\le 2\sum_{h=1}^H h\kappa_h\|\epsilon_h\|_{2,\nu_h}  \le 2\kappa H^2 \max_{1\le h \le H}\|\epsilon_h\|_{2,\nu_h}.
\end{equation}
Therefore, using the estimate \eqref{one_step_error} and the observation that
\begin{equation}
      \max_{\pi \in \mathcal{P}(\bA \cond \bS,H)}J_\nu(\pi) -  J_\nu(\hat{\pi}) \le     \|V_1^{*}- V_1^{\hat{\pi}}\|_{1,\nu_1},
\end{equation}
we finish our proof.

\subsection{Proof of Propositions \ref{Proposition_RKHS} and Related Discussion}\label{Appendix_proposition_2}
We directly verify Assumption \ref{main_assumption} (General) in the general setting of Remark \ref{rem_7}. Then proposition \ref{Proposition_RKHS} becomes a corollary when $\epsilon_r=\epsilon_p=0$. We first verify Assumption \ref{main_assumption} (General) \ref{ass11}. For any $\tilde{f} \in C(\bS)$ such that $0 \le \tilde{f} \le H$, we have
\begin{align}
    &\inf_{\|g\|_{\mathcal{H}_k} \le R}\| \mathcal{T}_h \tilde{f}(\,\cdot\,,a) - g\|_{2,\nu_{h,a}} \\
    =~&  \inf_{\|g\|_{\mathcal{H}_k} \le R}\| r_h(\,\cdot\,,a) + \int_{\bS} \tilde{f}(x')\rmd \bP_h(x'\cond \cdot\,,a) - g\|_{2,\nu_{h,a}} \\
    \le~& \inf_{\|g\|_{\mathcal{H}_k} \le K_r}\|g - r_h(\,\cdot\,,a)\|_{2,\nu_{h,a}} + \int_{\bS}\inf_{\|g\|_{\mathcal{H}_k} \le HK_p}\|g - \tilde{f}(x') p_h(\,\cdot\,,a,x')\|_{2,\nu_{h,a}} \rmd \rho_{h,a}(x')\notag\\
    \le~& \epsilon_r + H\int_{\bS}\inf_{\|g\|_{\mathcal{H}_k} \le K_p}\|g - p_h(\,\cdot\,,a,x')\|_{2,\nu_{h,a}}\rmd \rho_{h,a}(x') \notag \\
    \le~& \epsilon_f.
\end{align}
Therefore,
\begin{equation}
    \inf_{g \in \mathcal{F}_R}\|\mathcal{T}_h \tilde{f} - g\|_{2,\nu} \le \max_{a \in \bA} \inf_{\|g\|_{\mathcal{H}_k} \le R}\| \mathcal{T}_h \tilde{f}(\,\cdot\,,a) - g\|_{2,\nu_{h,a}} \le \epsilon_f.
\end{equation}
Noticing $\mathcal{T}_h^{*}f = \mathcal{T}_h \tilde{f}$ for any $f \in \mathcal{F}$ and corresponding $\tilde{f}$ defined by 
\begin{equation}
    \tilde{f}(x) = \mathcal{K}_{H-h} \max_{a'\in \bA} f(x,a'),
\end{equation}
 we are done with the first point. 
 
 Next, we verify Assumption \ref{main_assumption} (General) \ref{ass12} as follows.
\begin{align}
      \mathrm{Rad}_n(\mathcal{F}_r) &= \frac{1}{n}\EE \sup_{f \in\mathcal{F}_r} \sum_{i=1}^n\xi_i f(x_i,a_i) 
    =\frac{1}{n}\EE(\EE [\sup_{f \in\mathcal{F}_r} \sum_{i=1}^n\xi_i f(x_i,a_i)\cond a_1,\dots,a_n]) \notag \\
    &= \frac{1}{n}\EE(\EE [\sup_{f \in\mathcal{F}_r} \sum_{a \in \bA}\sum_{1\le i \le n, a_i = a}\xi_i f(x_i,a)\cond a_1,\dots,a_n]) \notag\\
    &\le \sum_{a \in \bA} \frac{1}{n}\EE(\EE[\sup_{\|f\|_{\mathcal{H}_k} \le r} \sum_{1 \le i \le n, a_i = a}\xi_i f(x_i)\cond a_1,\dots,a_n]) \notag\\
    & =  \sum_{a \in \bA} \frac{1}{n}\EE(\EE[\sup_{\|f\|_{\mathcal{H}_k} \le r}\langle f, \sum_{1 \le i \le n, a_i = a}\xi_i  k(x_i,\,\cdot\,)\rangle_{\mathcal{H}_k}\cond a_1,\dots,a_n]) \notag\\
    & =  \sum_{a \in \bA} \frac{1}{n}\EE(\EE[\|\sum_{1 \le i \le n,a_i = a}\xi_i k(x_i,\,\cdot\,)\|_{\mathcal{H}_k}\cond a_1,\dots,a_n]) \notag\\
    & =  \sum_{a \in \bA} \frac{1}{n}\EE(\EE[\sqrt{\sum_{1 \le i \le n,a_i = a}\sum_{1 \le j \le n,a_j = a}\xi_i\xi_j k(x_i,x_j)}\cond a_1,\dots,a_n]) \notag\\
    &\le \frac{\sqrt{|\bA|}}{n}\sqrt{\EE[\sum_{a \in \bA}\sum_{1\le i \le n,a_i = a}k(x_i,x_i)]} \le \sqrt{\frac{|\bA|}{n}\int_{\bS}k(x,x)\rmd \nu_h(x,a)}.
\end{align}

Now we give some discussions on the optimization problem~\eqref{optimization_problem}. Since $\mathcal{F}$ is an infinite dimensional space, it is hard to perform the optimization problem \eqref{optimization_problem} directly. We can prove that we can replace that problem with
\begin{equation}
      \overline{Q}_h = \argmin_{f \in \hat{\mathcal{F}}_h} \left\{\frac{1}{2n}[\sum_{i=1}^n |y_h^i -\mathcal{K}_{H-h+1} f(S_h^i,A_h^i)|^2] + \lambda\Lambda(f)\right\},
\end{equation}
and obtain the same result as Theorem \ref{main_theorem}. 
Here $\hat{\mathcal{F}}_h$ is a finite-dimensional space defined as
\begin{align}
    \hat{\mathcal{F}}_h = \{f \in C(\bS \times \bA), \exists\;b_i \in \bR, 1\le i \le n ,
    \text{such that } f(x,a) =\sum_{1 \le i \le n, A_h^i = a} b_i k(x,S_h^i)\}.
\end{align}
To see that, according to the proof of Theorem~\ref{main_theorem}, it is sufficient to prove that
\begin{equation}
    \min_{f \in \hat{\mathcal{F}}_h,\Lambda(f) \le R} L_n(f) \le \min_{f \in \mathcal{F}_R} L_n(f). 
\end{equation}
Let $g $ be the minimizer of $\min_{f \in \mathcal{F}_R} L_n(f)$.
Given
\begin{equation}
    \|g(\,\cdot\,,a)\|_{\mathcal{H}_k} \le R,
\end{equation}
there exist $\{g_a\}_{a \in \bA}$ such that
\begin{equation}
    g_a(x) = \sum_{i:A_h^i = a}b_i k(x,S_h^i),
\end{equation}
$\|g_a\|_{\mathcal{H}_k} \le R$ and $g_a(S_h^i) = g(S_h^i,a)$ \cite[Proposition~4.2]{paulsen2016introduction}. Let $\hat{g}(x,a) = g_a(x)$, then $g_a(x) \in \hat{\mathcal{F}}_h$, $L_n(\hat{g}) = L_n(g) = \min_{f \in \mathcal{F}_R} L_n(f)$ and $\Lambda(g) \le R$. Therefore
\begin{equation}
    \min_{f \in \hat{F}_h,\Lambda(f) \le R} L_n(f) \le \min_{f \in \mathcal{F}_R} L_n(f).
\end{equation}

Finally, if we go back to the setting of Propositions \ref{Proposition_RKHS}, we can drop the truncation operator to replace the optimization problem \eqref{optimization_problem} with
\begin{equation}
      \overline{Q}_h = \argmin_{f \in \mathcal{F}} \left\{\frac{1}{2n}[\sum_{i=1}^n |y_h^i - f(S_h^i,A_h^i)|^2] + \lambda\Lambda(f)\right\},
\end{equation}
and again replace $\mathcal{F}$ with $\hat{\mathcal{F}}_h$ using the same argument above. 
To see it, we notice that $\mathcal{T}_h^{*} Q_{h+1} \in \mathcal{F}$ and $\Lambda(\mathcal{T}_h^{*}Q_{h+1}) \le R$.
Then with probability at least $1-\frac{\delta}{2}$,
we have
\begin{align}
    &\quad L_n(\overline{Q}_h) + \lambda\Lambda(\overline{Q}_h) \notag \\
    &\le \frac{1}{2n}\sum_{i=1}^{n}|r_h(S_h^i,A_h^i) +V_{h+1}(\hat{S}^i_{h+1}) - \overline{Q}_h(S_h^i,A_h^i)|^2 + \lambda\Lambda(\overline{Q}_h)\notag\\
    &\le \frac{1}{2n}\sum_{i=1}^{n}|r_h(S_h^i,A_h^i) +V_{h+1}(\hat{S}^i_{h+1}) - (\mathcal{T}_h^{*}Q_{h+1})(S_h^i,A_h^i)|^2 + \lambda R \notag\\
    &= \frac{1}{2n}\sum_{i=1}^n\left|V_{h+1}(\hat{S}_{h+1}^i) - \EE[V_{h+1}(\hat{S}_{h+1}^i)\cond S_h^i,A_h^i]\right|^2 + \lambda R \notag\\
    &\le  \frac{1}{2}\EE_{(x,a)\sim \nu_h,x' \sim \bP_h(\,\cdot\cond x,a)}\left|V_{h+1}(x')-\EE[V_{h+1}(x')\cond x,a] \right|^2 + H^2 \sqrt{\frac{2\ln(2/\delta)}{n}} + \lambda R,
\end{align}
where the last inequality follows the Hoeffding's inequality \cite[Theorem~2.2.6]{vershynin2018high}.
Using the above inequality to replace inequality \eqref{Q-Estimation} and \eqref{decomposition_min} and following the proof of Theorem \ref{main_theorem} gives us a similar estimate.

\subsection{Proof of Proposition \ref{Proposition_Barron}}
In the setting of Remark \ref{rem_10}, with an argument similar to that in the proof of Proposition \ref{Proposition_RKHS}, we can prove that
\begin{equation}
    \inf_{\|g\|_\mathcal{B} \le \frac{R}{2}}\| (\mathcal{T}_h^{*}f)(\,\cdot\,,a) - g\|_{2,\nu_{h,a}} \le \epsilon_r + H\epsilon_p,
\end{equation}
where $R=2(B_r + HB_p)$.
Moreover, following Theorem 4 in \cite{weinan2019barron}, for any $g \in \mathcal{B}$ so that $\|g\|_\mathcal{B} \le \frac{R}{2}$, there exists $b_1,\dots,b_m \in \bR$ and $\omega_1,\dots,\omega_m \in \bR^d$, such that
\begin{equation}
    \|g(x) - \frac{1}{m}\sum_{i=1}^m b_i\sigma(\omega_i\cdot x) \|_{2,\nu_{h,a}} \le Rm^{-\frac{1}{2}}
\end{equation}
and
\begin{equation}
    \frac{1}{m}\sum_{i=1}^m|b_i|\|\omega_i\| \le R.
\end{equation}

We then obtain
\begin{equation}
    \inf_{g \in \mathcal{F}_R }\|(\mathcal{T}_h^{*}f) - g\|_{2,\nu_{h}} \le \epsilon_r + H\epsilon_p + R m^{-\frac{1}{2}} = \epsilon_f,
\end{equation}
for any $f \in \mathcal{F}$.

We next verify Assumption \ref{main_assumption} (General) \ref{ass12}. Given $n' \in \mathbb{N}^{+}$ and $x_1,\dots,x_{n'} \in \mathbb{S}^{d-1}$, we have
\begin{align}
   &\EE \sup_{\frac{1}{m}\sum_{j=1}^m|b_j|\|\omega_j\| \le r} \sum_{i=1}^{n'}\frac{\xi_i}{m}  \sum_{j=1}^m\xi_i b_j \sigma(\omega_j \cdot x_i) \notag\\
    =\,&r \EE \sup_{\sum_{j=1}^m|b_j|\|\omega_j\|\le 1}\sum_{i=1}^{n'}\sum_{j=1}^m\xi_i |b_j|\|\omega_j\|\sigma(\frac{\omega_j}{\|\omega_j\|} \cdot x_i) \notag\\
    =\,&r\EE \sup_{\sum_{j=1}^m|b_j|\|\omega_j\|\le 1}\sum_{j=1}^m|b_j|\|\omega_j\|\sum_{i=1}^{n'}\xi_i \sigma(\frac{\omega_j}{\|\omega_j\|} \cdot x_i) \notag\\
    =\,& r\EE\sup_{\|\omega\| = 1}|\sum_{i=1}^{n'}\xi_i\sigma(\omega \cdot x_i)| \notag\\
    \le\,& r\EE[ \sup_{\|\omega\| = 1}\sum_{i=1}^{n'} \xi_i \sigma(\omega \cdot x_i) +\sup_{\|\omega\| = 1}-\sum_{i=1}^{n'} \xi_i \sigma(\omega \cdot x_i)]\notag\\
    =\,& 2r\EE \sup_{\|\omega\| = 1}\sum_{i=1}^{n'} \xi_i \sigma(\omega \cdot x_i).
\end{align}
Using Lemma 26.9 in \cite{shalev2014understanding}, we have
\begin{equation}
   \EE \sup_{\|\omega\| = 1}\sum_{i=1}^{n'} \xi_i \sigma(\omega \cdot x_i) \le\EE \sup_{\|\omega\| = 1}\sum_{i=1}^{n'} \xi_i(\omega \cdot x_i) = \EE\|\sum_{i=1}^{n'}\xi_i x_i\| \le \sqrt{\EE\|\sum_{i=1}^{n'}\xi_i x_i\|^2} = \sqrt{n'}.
\end{equation}
Now we are ready to use an argument similar to that in the proof of Proposition~\ref{Proposition_RKHS} to obtain
\begin{align}
    \mathrm{Rad}_n(\mathcal{F}_r) &\le \sum_{a \in \bA} \frac{1}{n}\EE(\EE[\sup_{\frac{1}{m}\sum_{j=1}^m|b_j|\|\omega_j\| \le r}\sum_{1 \le i \le n,a_i = a}\frac{\xi_i}{m}\sum_{j=1}^m \xi_i b_j\sigma(\omega_j\cdot x_i)\cond a_1,\dots,a_n])\notag\\
    &\le\frac{2r}{n} \sum_{a \in \bA}\EE(\EE[\sqrt{|\{i,a_i = a\}|}\cond a_1,\dots,a_n]) \le \frac{2r\sqrt{|\bA|}}{\sqrt{n}}.
\end{align}

\section{Sample Complexity of $L^2$ and $L^\infty$ Estimates in RKHS and Barron Space}
\label{section_sample}
In this section we analyze the lower bound of the sample complexity of $L^2$ and $L^\infty$ estimates in RKHS corresponding to different kernels and the Barron space. The analysis of $L^\infty$ estimate reveals the encountered curse of dimensionality and the analysis of $L^2$ estimate justifies our main result in terms of $n$ is optimal in asymptotics.
Given a positive kernel $k$ on $\bS \times \bS$ and a probability distributions $\pi$ on $\bS$, we define a linear operator $K: L^2(\pi)\rightarrow L^2(\pi)$:
\begin{equation}
    (K f)(x) = \int_{\bS}k(x,x')f(x')\rmd \pi(x'),
\end{equation}
assuming that $\int_{\bS} k(x,x) \rmd \pi(x) < + \infty$.  By Mercer's Theorem \cite{steinwart2008support}, $K$ has positive nonincreasing eigenvalues $\{\lambda_l\}_{l=1}^{\infty}$ and corresponding eigenfunctions $\{\psi_l\}_{l=1}^{\infty}$ that are orthonormal on $L^2(\pi)$ and
\begin{equation}
    k(x,x') = \sum_{l=1}^{\infty}\lambda_s \psi_l(x)\psi_l(x').
\end{equation}
In addition,
\begin{equation}
    \mathcal{H}_k = \{f \in L^2(\pi): \sum_{l=1}^{\infty}\frac{\langle f,\psi_l\rangle_{L^2(\pi)}^2}{\lambda_l} < +\infty\},
\end{equation}
and
\begin{equation}
    \langle f , g\rangle_{\mathcal{H}_k} = \sum_{l=1}^{\infty}\frac{\langle f,\psi_l \rangle_{L^2(\pi)}\langle g,\psi_l \rangle_{L^2(\pi)}}{\lambda_l}.
\end{equation}
We will then use $\mathcal{H}_{k,1}$ to denotes the unit ball in $\mathcal{H}_k$, which admits the following representation:
\begin{equation}
    \mathcal{H}_{k,1} = \{f = \sum_{l=1}^{\infty} b_l\psi_l, \; \sum_{l=1}^{\infty} \frac{b_l^2}{\lambda_l} \le 1 \}.
\end{equation}

Let $\bS = \mathbb{S}^{d-1}$ and $\pi$ be the uniform distribution on $\mathbb{S}^{d-1}$.
We are interested in the sample complexity of $L^2$ and $L^\infty$ estimates in the RKHS corresponding to the Laplacian kernel/neural tangent kernel, and the Barron space. To this end, we need to analyze the eigenvalues of the following operators
\begin{equation}
    (K_i f)(x) = \int_{\mathbb{S}^{d-1}}k_i(x,x')f(x')\rmd \pi(x'), \quad i = 1,2,3,
\end{equation}
\begin{equation}
    k_i(x,x') = \begin{cases} k_{Lap}(x,x') = \exp(-\|x - x'\|) &\text{ when } i = 1 \\
    k_{NTK}(x,x') = \EE_{\omega \sim \pi}( x \cdot x')\sigma'(\omega \cdot x)\sigma'(\omega \cdot x')  &\text{ when } i =2 \\
    k_\pi(x,x') = \EE_{\omega \sim \pi}\sigma(\omega \cdot x)\sigma(\omega\cdot x') &\text{ when } i =3
    \end{cases}
\end{equation}
Here, we only consider the NTK kernel for two-layer neural networks without bias term. Similar results can be obtained for  deep fully connected neural network with zero-initialization bias term \cite{geifman2020similarity,chen2020deep}.
$\{k_i\}_{i=1}^3$ admit similar Mercer decompositions
\begin{equation}
    k_i(x,x') = \sum_{l=0}^\infty \mu_{l,i}\sum_{j=1}^{N(d,l)}Y_{l,j}(x)Y_{l,j}(x'), 
\end{equation}
where $Y_{l,j}$, $j = 1,\dots, N(d,l)$ are spherical harmonics of degree $l$, 
\begin{equation}
    N(d,l) = (2l+d-2) \frac{(l+d -3)!}{(d-2)!l!} \sim l^{d-2}
\end{equation}
for $l \ge 1$, $N(d,0) = 1$ and 
\begin{enumerate}
    \item $\mu_{l,1} \sim l^{-d}$ for $l \ge 1$ and $\mu_{0,1} > 0$ (see \cite{geifman2020similarity});
    \item $\mu_{2l,2} \sim l^{-d}$, $\mu_{2l+1,2} = 0$ for $l \ge 1$ and $\mu_{0,2},\mu_{1,2} > 0$ (see \cite{bietti2019inductive});
    \item $\mu_{2l,3} \sim l^{-d-2}$, $\mu_{2l+1,3} = 0$ for $l \ge 1$ and $\mu_{0,3},\mu_{1,3} > 0$ (see \cite{bach2017breaking}),
\end{enumerate}
where $f(l) \sim g(l)$ means that $f(l) \le O(g(l))$ and $g(l) \le O(f(l))$.
Therefore, let $\{\lambda_{l,i}\}_{l=1}^\infty$ be the nonincreasing eigenvalues of $K_i$ for $i = 1,2,3$, we have
\begin{equation}
    \lambda_{l,1} = \mu_{p,1} \text{ if }\sum_{j = 0}^{p-1}N(d,j) + 1 \le l \le \sum_{j=0}^p N(d,j),
\end{equation}
and
\begin{equation}
    \lambda_{l,i} = \mu_{2p,i} \text{ if } d  + 2 + \sum_{j=0}^{p-1}N(d,2j)\le l \le  d+1 + \sum_{j=0}^{p} N(d,2j) \text{ and } p \ge 1,
\end{equation}
for $i = 2,3$. 
In summary,
\begin{equation}\label{eigenvalue_decay}
    \lambda_{l,i}\sim \begin{cases}
      l^{-\frac{d}{d-1}} &\text{ when }i=1,2;\\
      l^{-\frac{d+2}{d-1}} &\text{ when }i=3.
    \end{cases}
\end{equation}

\bigskip
\noindent\textbf{Sample Complexity of $L^\infty$ Estimation}: Fix a target function $f^{*} \in \mathcal{H}_{k,1}$ and $x_1,\dots,x_n \in \bS$. With the knowledge of $f^{*}(x_1),\dots,f^{*}(x_n)$, we want to recover $f^{*}$ as $f$ so that $\|f^{*} - f\|_\infty$ is as small as possible. Let
\begin{equation}
    \mathbb{T}_n(x_1,\dots,x_n) = \{ T: \mathcal{H}_{k,1} \rightarrow C(\bS), \text{ there exists } \Psi: \bR^d \rightarrow C(\bS) \text{ such that }(Tf) = \Psi(f(x_1),\dots,f(x_n))\},
\end{equation}
which contains all estimators on $\mathcal{H}_{k,1}$ only depending on $f(x_1),\dots,f(x_n)$.
It is shown in \cite{kuo2008multivariate} that the worst $L^\infty$ error over is $\mathcal{H}_{k,1}$ 
\begin{equation}
    \sup_{f \in \mathcal{H}_{k,1}} \|f - Tf\|_\infty \ge (\sum_{l = n+1}^\infty\lambda_l)^{\frac{1}{2}},
\end{equation}
for any $x_1,\dots,x_n \in \bS$ and $T \in \mathbb{T}_n(x_1,\dots,x_n)$.
We shall remark that the above result is based on the fact that
for any $x_1,\dots,x_n$, there exists function $\hat{f} \in \mathcal{H}_{k,1}$ such that $\hat{f}(x_1) = \dots = \hat{f}(x_n) = 0$ and $\|\hat{f}\|_\infty \ge (\sum_{l=n+1}^\infty \lambda_l)^{\frac{1}{2}}$ and hence any $T \in \mathbb{T}_n(x_1,\dots,x_n)$ can not approximate $\hat{f}$ and $-\hat{f}$ less than $(\sum_{l=n+1}^\infty \lambda_l)^{\frac{1}{2}}$ simultaneously. Therefore, neither the adaptive sampling nor the non-deterministic estimator can break this lower bound. 

Applying the above argument to $\mathcal{H}_{k_i,1}$ with inequality \eqref{eigenvalue_decay}, we can obtain that there exists a constant $c > 0$ such that
\begin{equation}
    \sup_{f \in \mathcal{H}_{{k_i},1}}\|f - Tf\|_\infty \ge \begin{cases}c n^{-\frac{1}{d-1}} & \text{ when } i = 1,2;\\
    cn^{-\frac{3}{d-1}} & \text{ when } i = 3,
    \end{cases}
\end{equation}
for any estimator $T$ depending only on function values of $n$ points, which means that  $L^\infty$ estimates in the unit balls of $\mathcal{H}_{k_{Lap}}$, $\mathcal{H}_{k_{NTK}}$ and $\mathcal{H}_{k_\pi}$  suffer from the curse of dimensionality in regard to sample complexity. Since the unit ball of the Barron space $\mathcal{B}$ contains the unit ball of $\mathcal{H}_{k_\pi}$ (see Theorem 3 in \cite{weinan2019barron}), 
the lower bound above also applies to the Barron space.

We shall remark that if $\bA$ is not finite but in high dimensions, with the similar assumption of Assumption \ref{assumption_rkhs}~(RKHS), it can be shown that $Q_h^{*}(x,\cdot)$ is in a certain RKHS for any $x \in \bS$ with uniform bounded norm. The following important issue is to find the maximum of a function in RKHS, which admits the same lower bound due to the existence of $\hat{f}$. 

\bigskip
\noindent\textbf{Sample Complexity of $L^2$ Estimation}:  Consider the nonparametric regression model
\begin{equation}
    y_i = f^{*}(x_i) + \epsilon_i,\quad i = 1,\dots,n,
\end{equation}
where $f^{*} \in \mathcal{H}_{k,1}$, $\{x_i\}_{i=1}^n$ are i.i.d. sampled from $\pi$ and $\{\epsilon_i\}_{i=1}^n$ are i.i.d. sampled from the standard normal distribution $\mathcal{ N}(0,1)$. Let $\hat{u}:\bR^n \rightarrow C(\bS)$ be a measurable mapping. By Theorem 6 and Example 6.1 in \cite{yang1999information}, if $\lambda_l \sim l^{-\alpha}$ for constant $\alpha > 1$, then
\begin{equation}
    \inf_{\hat{u}}\sup_{f^{*} \in \mathcal{H}_{k,1}}\EE\|\hat{u}(y_1,\dots,y_n) - f^{*}\|_{p,\pi} \sim n^{-\frac{\alpha}{2(\alpha+1)}},
\end{equation}
for any $p \in [1,2]$.  Applying inequality \eqref{eigenvalue_decay}, we know that there exists a constant $c > 0$ such that
\begin{equation}
    \sup_{f^{*} \in \mathcal{H}_{k_i,1}}\|\hat{u}(y_1,\dots,y_n) - f^{*}\|_{p,\pi} \ge  \begin{cases} cn^{-\frac{d}{4d-2}} &\text{ when }i = 1,2;\\
    cn^{-\frac{d+2}{4d+2}} &\text{ when }i =3,
    \end{cases}
\end{equation}
for any measurable $\hat{u}: \bR^n \rightarrow C(\mathbb{S}^{d-1})$ and $p \in [1,2]$. This results shows that the convergences rate of $L^2$ estimate with noise in the unit balls of $\mathcal{H}_{k_{Lap}}$, $\mathcal{H}_{k_{NTK}}$, $\mathcal{H}_{k_\pi}$ and Barron Space $\mathcal{B}$ is near $n^{-\frac{1}{4}}$ when $d$ is large. Hence, the rate of $n$ in our main result is optimal in asymptotics.
We also remark that this statistical lower bound does not violate the convergence rate of $n^{-\frac{1}{2}}$ in previous works~\cite{azar2012sample, azar2017minimax, jin2020provably,wang2019optimism,yang2019sample,chen2019information} since the eigenvalues can be viewed as exponentially decaying in those settings.

\section{Conclusion and Discussion}
This work analyzes reinforcement learning in high dimensions with kernel and neural network approximation and establishes an $\tilde{O}(H^3|\bA|^{\frac{1}{4}}n^{-\frac{1}{4}})$ bound for the optimal policy in the fitted Q-iteration algorithm. We note that for function spaces like Barron space and many popular cases of RKHS, the convergence rate $n^{-\frac{1}{4}}$ is close to the statistical lower bound in high dimensions, as discussed in Section \ref{section_sample}. Replacing Rademacher complexity by local Rademacher complexity \cite{bartlett2005local,cortes2013learning} in the analysis can improve the convergence rate with respect to $n$, particularly for kernels with fast decaying eigenvalues.
This would basically fill the gap with the statistical lower bound shown in Section \ref{section_sample} (see \cite{cortes2013learning} for a detailed discussion).

There are still many open problems related to the topic discussed here. Firstly, although the convergence rate with respect to $n$ is near-optimal, it is not clear whether the convergence rate with respect to $H$ and $|\bA|$ can be improved. Secondly, Assumption \ref{assumption_rkhs}~(RKHS) and Assumption \ref{Assumption_Barron}~(Barron) are only sufficient conditions of Assumption \ref{main_assumption}~(General) in their own settings. To what extent these assumptions can be relaxed remains unclear. Thirdly,  algorithms studied here require choosing suitable $\{\nu_h\}_{h=1}^H$ with small $\kappa$. Although we suggested some ideas about  choosing these sampling distributions, 
 how to choose $\{\nu_h\}_{h=1}^H$ from the prior knowledge of MDP and the  data collected  is still an interesting issue.
 Fourthly, our assumptions on MDP exclude non-trivial deterministic MDPs. It is unclear how to extend the current framework to the setting of deterministic or near-deterministic MDPs.
Finally and perhaps most importantly, in the episodic reinforcement learning environment, 
it is still challenging to combine efficient exploration and general function approximation to design reliable algorithms in high-dimensional space.
We hope to explore these directions in future work.
\bibliographystyle{plain}
\bibliography{ref}

\begin{thebibliography}{10}

\bibitem{aronszajn1950theory}
Nachman Aronszajn.
\newblock Theory of reproducing kernels.
\newblock {\em Transactions of the American mathematical society},
  68(3):337--404, 1950.

\bibitem{arora2019fine}
Sanjeev Arora, Simon Du, Wei Hu, Zhiyuan Li, and Ruosong Wang.
\newblock Fine-grained analysis of optimization and generalization for
  overparameterized two-layer neural networks.
\newblock In {\em International Conference on Machine Learning}, pages
  322--332. PMLR, 2019.

\bibitem{azar2012sample}
Mohammad~Gheshlaghi Azar, R\'{e}mi Munos, and Hilbert~J. Kappen.
\newblock On the sample complexity of reinforcement learning with a generative
  model.
\newblock In {\em Proceedings of the 29th International Coference on
  International Conference on Machine Learning}, page 1707–1714, 2012.

\bibitem{azar2017minimax}
Mohammad~Gheshlaghi Azar, Ian Osband, and R{\'e}mi Munos.
\newblock Minimax regret bounds for reinforcement learning.
\newblock In {\em International Conference on Machine Learning}, pages
  263--272. PMLR, 2017.

\bibitem{bach2017breaking}
Francis Bach.
\newblock Breaking the curse of dimensionality with convex neural networks.
\newblock {\em The Journal of Machine Learning Research}, 18(1):629--681, 2017.

\bibitem{bartlett2005local}
Peter~L Bartlett, Olivier Bousquet, and Shahar Mendelson.
\newblock Local rademacher complexities.
\newblock {\em The Annals of Statistics}, 33(4):1497--1537, 2005.

\bibitem{bartlett2002rademacher}
Peter~L Bartlett and Shahar Mendelson.
\newblock Rademacher and gaussian complexities: Risk bounds and structural
  results.
\newblock {\em Journal of Machine Learning Research}, 3(Nov):463--482, 2002.

\bibitem{bietti2019inductive}
Alberto Bietti and Julien Mairal.
\newblock On the inductive bias of neural tangent kernels.
\newblock {\em arXiv preprint arXiv:1905.12173}, 2019.

\bibitem{bradtke1996linear}
Steven~J Bradtke and Andrew~G Barto.
\newblock Linear least-squares algorithms for temporal difference learning.
\newblock {\em Machine learning}, 22(1):33--57, 1996.

\bibitem{bubeck2012regret}
S{\'{e}}bastien Bubeck.
\newblock Regret analysis of stochastic and nonstochastic multi-armed bandit
  problems.
\newblock {\em Foundations and Trends{\textregistered} in Machine Learning},
  5(1):1--122, 2012.

\bibitem{cai2019provably}
Qi~Cai, Zhuoran Yang, Chi Jin, and Zhaoran Wang.
\newblock Provably efficient exploration in policy optimization.
\newblock In {\em International Conference on Machine Learning}, pages
  1283--1294. PMLR, 2020.

\bibitem{cai2019neural}
Qi~Cai, Zhuoran Yang, Jason~D Lee, and Zhaoran Wang.
\newblock Neural temporal-difference learning converges to global optima.
\newblock {\em Advances in Neural Information Processing Systems}, 32, 2019.

\bibitem{chen2019information}
Jinglin Chen and Nan Jiang.
\newblock Information-theoretic considerations in batch reinforcement learning.
\newblock In {\em International Conference on Machine Learning}, pages
  1042--1051. PMLR, 2019.

\bibitem{chen2020deep}
Lin Chen and Sheng Xu.
\newblock Deep neural tangent kernel and laplace kernel have the same {RKHS}.
\newblock {\em arXiv preprint arXiv:2009.10683}, 2020.

\bibitem{cortes2013learning}
Corinna Cortes, Marius Kloft, and Mehryar Mohri.
\newblock Learning kernels using local rademacher complexity.
\newblock In {\em Advances in Neural Information Processing Systems},
  volume~26, pages 2760--2768, 2013.

\bibitem{cortes2010generalization}
Corinna Cortes, Mehryar Mohri, and Afshin Rostamizadeh.
\newblock Generalization bounds for learning kernels.
\newblock In {\em 27th International Conference on Machine Learning, ICML
  2010}, pages 247--254, 2010.

\bibitem{dann2017unifying}
Christoph Dann, Tor Lattimore, and Emma Brunskill.
\newblock Unifying {PAC} and regret: Uniform {PAC} bounds for episodic
  reinforcement learning.
\newblock In {\em Advances in Neural Information Processing Systems},
  volume~30, pages 5713--5723, 2017.

\bibitem{domingues2020regret}
Omar~Darwiche Domingues, Pierre M{\'e}nard, Matteo Pirotta, Emilie Kaufmann,
  and Michal Valko.
\newblock Regret bounds for kernel-based reinforcement learning.
\newblock {\em arXiv preprint arXiv:2004.05599}, 2020.

\bibitem{du2019provably}
Simon~S Du, Yuping Luo, Ruosong Wang, and Hanrui Zhang.
\newblock Provably efficient $ q $-learning with function approximation via
  distribution shift error checking oracle.
\newblock {\em arXiv preprint arXiv:1906.06321}, 2019.

\bibitem{duan2016benchmarking}
Yan Duan, Xi~Chen, Rein Houthooft, John Schulman, and Pieter Abbeel.
\newblock Benchmarking deep reinforcement learning for continuous control.
\newblock In {\em International Conference on Machine Learning}, pages
  1329--1338. PMLR, 2016.

\bibitem{weinan2020towards}
Weinan E, Chao Ma, Stephan Wojtowytsch, and Lei Wu.
\newblock Towards a mathematical understanding of neural network-based machine
  learning: What we know and what we don’t.
\newblock {\em arXiv preprint arXiv:2009.10713}, 2020.

\bibitem{weinan2019barron}
Weinan E, Chao Ma, and Lei Wu.
\newblock Barron spaces and the compositional function spaces for neural
  network models.
\newblock {\em arXiv preprint arXiv:1906.08039}, 2019.

\bibitem{ma2019comparative}
Weinan E, Chao Ma, and Lei Wu.
\newblock A comparative analysis of the optimization and generalization
  property of two-layer neural network and random feature models under gradient
  descent dynamics.
\newblock {\em arXiv preprint arXiv:1904.04326}, 2019.

\bibitem{ma2019priori}
Weinan E, Chao Ma, and Lei Wu.
\newblock A priori estimates of the population risk for two-layer neural
  networks.
\newblock {\em Communications in Mathematical Sciences}, 17(5):1407--1425,
  2019.

\bibitem{weinan2021kolmogorov}
Weinan E and Stephan Wojtowytsch.
\newblock Kolmogorov width decay and poor approximators in machine learning:
  Shallow neural networks, random feature models and neural tangent kernels.
\newblock {\em Research in the Mathematical Sciences}, 8(1):1--28, 2021.

\bibitem{ernst2005tree}
Damien Ernst, Pierre Geurts, and Louis Wehenkel.
\newblock Tree-based batch mode reinforcement learning.
\newblock {\em Journal of Machine Learning Research}, 6(Apr):503--556, 2005.

\bibitem{fan2020theoretical}
Jianqing Fan, Zhaoran Wang, Yuchen Xie, and Zhuoran Yang.
\newblock A theoretical analysis of deep {Q}-learning.
\newblock In {\em Learning for Dynamics and Control}, pages 486--489. PMLR,
  2020.

\bibitem{farahmand2016regularized}
Amir-massoud Farahmand, Mohammad Ghavamzadeh, Csaba Szepesv{\'a}ri, and Shie
  Mannor.
\newblock Regularized policy iteration with nonparametric function spaces.
\newblock {\em The Journal of Machine Learning Research}, 17(1):4809--4874,
  2016.

\bibitem{farahmand2010error}
Amir-massoud Farahmand, Csaba Szepesv{\'a}ri, and R{\'e}mi Munos.
\newblock Error propagation for approximate policy and value iteration.
\newblock {\em Advances in Neural Information Processing Systems}, 23:568--576,
  2010.

\bibitem{geifman2020similarity}
Amnon Geifman, Abhay Yadav, Yoni Kasten, Meirav Galun, David Jacobs, and Ronen
  Basri.
\newblock On the similarity between the laplace and neural tangent kernels.
\newblock {\em arXiv preprint arXiv:2007.01580}, 2020.

\bibitem{jacot2018neural}
Arthur Jacot, Franck Gabriel, and Cl{\'e}ment Hongler.
\newblock Neural tangent kernel: Convergence and generalization in neural
  networks.
\newblock In {\em Advances in Neural Information Processing Systems}, pages
  8571--8580, 2018.

\bibitem{jaksch2010near}
Thomas Jaksch, Ronald Ortner, and Peter Auer.
\newblock Near-optimal regret bounds for reinforcement learning.
\newblock {\em Journal of Machine Learning Research}, 11(4), 2010.

\bibitem{jin2018q}
Chi Jin, Zeyuan Allen-Zhu, Sebastien Bubeck, and Michael~I Jordan.
\newblock Is {Q}-learning provably efficient?
\newblock In {\em Advances in Neural Information Processing Systems},
  volume~31, pages 4863--4873, 2018.

\bibitem{jin2020provably}
Chi Jin, Zhuoran Yang, Zhaoran Wang, and Michael~I Jordan.
\newblock Provably efficient reinforcement learning with linear function
  approximation.
\newblock In {\em Conference on Learning Theory}, pages 2137--2143, 2020.

\bibitem{kakade2003sample}
Sham~Machandranath Kakade.
\newblock {\em On the sample complexity of reinforcement learning}.
\newblock PhD thesis, UCL (University College London), 2003.

\bibitem{kuo2008multivariate}
Frances~Y Kuo, Grzegorz~W Wasilkowski, and Henryk Wo{\'z}niakowski.
\newblock Multivariate ${L}_\infty$ approximation in the worst case setting
  over reproducing kernel {H}ilbert spaces.
\newblock {\em Journal of approximation theory}, 152(2):135--160, 2008.

\bibitem{lazaric2016analysis}
Alessandro Lazaric, Mohammad Ghavamzadeh, and R{\'e}mi Munos.
\newblock Analysis of classification-based policy iteration algorithms.
\newblock {\em The Journal of Machine Learning Research}, 17(1):583--612, 2016.

\bibitem{liu2019neural}
Boyi Liu, Qi~Cai, Zhuoran Yang, and Zhaoran Wang.
\newblock Neural trust region/proximal policy optimization attains globally
  optimal policy.
\newblock {\em Advances in Neural Information Processing Systems},
  32:10565--10576, 2019.

\bibitem{melo2007q}
Francisco~S Melo and M~Isabel Ribeiro.
\newblock Q-learning with linear function approximation.
\newblock In {\em International Conference on Computational Learning Theory},
  pages 308--322. Springer, 2007.

\bibitem{mnih2013playing}
Volodymyr Mnih, Koray Kavukcuoglu, David Silver, Alex Graves, Ioannis
  Antonoglou, Daan Wierstra, and Martin Riedmiller.
\newblock Playing atari with deep reinforcement learning.
\newblock {\em arXiv preprint arXiv:1312.5602}, 2013.

\bibitem{munos2008finite}
R{\'e}mi Munos and Csaba Szepesv{\'a}ri.
\newblock Finite-time bounds for fitted value iteration.
\newblock {\em Journal of Machine Learning Research}, 9:815--857, 2008.

\bibitem{neyshabur2018towards}
Behnam Neyshabur, Zhiyuan Li, Srinadh Bhojanapalli, Yann LeCun, and Nathan
  Srebro.
\newblock Towards understanding the role of over-parametrization in
  generalization of neural networks.
\newblock {\em arXiv preprint arXiv:1805.12076}, 2018.

\bibitem{osband2016generalization}
Ian Osband, Benjamin Van~Roy, and Zheng Wen.
\newblock Generalization and exploration via randomized value functions.
\newblock In {\em International Conference on Machine Learning}, pages
  2377--2386. PMLR, 2016.

\bibitem{paulsen2016introduction}
Vern~I Paulsen and Mrinal Raghupathi.
\newblock {\em An introduction to the theory of reproducing kernel Hilbert
  spaces}, volume 152.
\newblock Cambridge University Press, 2016.

\bibitem{puterman2014markov}
Martin~L Puterman.
\newblock {\em Markov decision processes: discrete stochastic dynamic
  programming}.
\newblock John Wiley \& Sons, 2014.

\bibitem{riedmiller2005neural}
Martin Riedmiller.
\newblock Neural fitted {Q} iteration--first experiences with a data efficient
  neural reinforcement learning method.
\newblock In {\em European Conference on Machine Learning}, pages 317--328.
  Springer, 2005.

\bibitem{scherrer2015approximate}
Bruno Scherrer, Mohammad Ghavamzadeh, Victor Gabillon, Boris Lesner, and
  Matthieu Geist.
\newblock Approximate modified policy iteration and its application to the game
  of tetris.
\newblock {\em Journal of Machine Learning. Research}, 16:1629--1676, 2015.

\bibitem{shalev2014understanding}
Shai Shalev-Shwartz and Shai Ben-David.
\newblock {\em Understanding machine learning: From theory to algorithms}.
\newblock Cambridge University Press, 2014.

\bibitem{silver2016mastering}
David Silver, Aja Huang, Chris~J Maddison, Arthur Guez, Laurent Sifre, George
  Van Den~Driessche, Julian Schrittwieser, Ioannis Antonoglou, Veda
  Panneershelvam, Marc Lanctot, et~al.
\newblock Mastering the game of go with deep neural networks and tree search.
\newblock {\em nature}, 529(7587):484--489, 2016.

\bibitem{steinwart2008support}
Ingo Steinwart and Andreas Christmann.
\newblock {\em Support vector machines}.
\newblock Springer Science \& Business Media, 2008.

\bibitem{szepesvari2010algorithms}
Csaba Szepesv{\'a}ri.
\newblock Algorithms for reinforcement learning.
\newblock {\em Synthesis lectures on artificial intelligence and machine
  learning}, 4(1):1--103, 2010.

\bibitem{vershynin2018high}
Roman Vershynin.
\newblock {\em High-dimensional probability: An introduction with applications
  in data science}, volume~47.
\newblock Cambridge University Press, 2018.

\bibitem{wang2019neural}
Lingxiao Wang, Qi~Cai, Zhuoran Yang, and Zhaoran Wang.
\newblock Neural policy gradient methods: Global optimality and rates of
  convergence.
\newblock {\em arXiv preprint arXiv:1909.01150}, 2019.

\bibitem{wang2020provably}
Ruosong Wang, Ruslan Salakhutdinov, and Lin~F Yang.
\newblock Provably efficient reinforcement learning with general value function
  approximation.
\newblock {\em arXiv preprint arXiv:2005.10804}, 2020.

\bibitem{wang2019optimism}
Yining Wang, Ruosong Wang, Simon~S Du, and Akshay Krishnamurthy.
\newblock Optimism in reinforcement learning with generalized linear function
  approximation.
\newblock {\em arXiv preprint arXiv:1912.04136}, 2019.

\bibitem{wen2013efficient}
Zheng Wen and Benjamin Van~Roy.
\newblock Efficient exploration and value function generalization in
  deterministic systems.
\newblock {\em Advances in Neural Information Processing Systems}, 2013.

\bibitem{yang2019sample}
Lin Yang and Mengdi Wang.
\newblock Sample-optimal parametric q-learning using linearly additive
  features.
\newblock In {\em International Conference on Machine Learning}, pages
  6995--7004. PMLR, 2019.

\bibitem{yang2020reinforcement}
Lin~F Yang and Mengdi Wang.
\newblock Reinforcement learning in feature space: Matrix bandit, kernels, and
  regret bound.
\newblock In {\em International Conference on Machine Learning}, pages
  10746--10756. PMLR, 2020.

\bibitem{yang1999information}
Yuhong Yang and Andrew Barron.
\newblock Information-theoretic determination of minimax rates of convergence.
\newblock {\em Annals of Statistics}, pages 1564--1599, 1999.

\bibitem{yang2020provably}
Zhuoran Yang, Chi Jin, Zhaoran Wang, Mengdi Wang, and Michael Jordan.
\newblock Provably efficient reinforcement learning with kernel and neural
  function approximations.
\newblock {\em Advances in Neural Information Processing Systems}, 33, 2020.

\bibitem{yang2020function}
Zhuoran Yang, Chi Jin, Zhaoran Wang, Mengdi Wang, and Michael~I Jordan.
\newblock On function appproximation in reinforcement learning: Optimisim in
  the face of large state spaces.
\newblock {\em arXiv preprint arXiv:2011.04622}, 2020.

\bibitem{zanette2020frequentist}
Andrea Zanette, David Brandfonbrener, Emma Brunskill, Matteo Pirotta, and
  Alessandro Lazaric.
\newblock Frequentist regret bounds for randomized least-squares value
  iteration.
\newblock In {\em International Conference on Artificial Intelligence and
  Statistics}, pages 1954--1964. PMLR, 2020.

\bibitem{zhou2002covering}
Ding-Xuan Zhou.
\newblock The covering number in learning theory.
\newblock {\em Journal of Complexity}, 18(3):739--767, 2002.

\bibitem{zhou2003capacity}
Ding-Xuan Zhou.
\newblock Capacity of reproducing kernel spaces in learning theory.
\newblock {\em IEEE Transactions on Information Theory}, 49(7):1743--1752,
  2003.

\end{thebibliography}
\end{document}